\documentclass[10pt,twocolumn,letterpaper]{article}

\usepackage{iccv}
\usepackage{times}
\usepackage{epsfig}
\usepackage{graphicx}
\usepackage{amsmath}
\usepackage{amssymb}

\usepackage{bbm}
\usepackage{booktabs}
\usepackage{multirow}
\usepackage[table]{xcolor}
\usepackage{subfig}
\usepackage{bbding}
\usepackage[accsupp]{axessibility}

\DeclareMathOperator*{\argmin}{argmin}

\usepackage[pagebackref=true,breaklinks=true,colorlinks,bookmarks=false]{hyperref}

\iccvfinalcopy 

\def\iccvPaperID{****} 

\ificcvfinal\fi

\begin{document}

\title{Audio-Visual Class-Incremental Learning}


\author{Weiguo Pian\textsuperscript{\rm 1}\footnotemark[2] , Shentong Mo\textsuperscript{\rm 2}\footnotemark[2] , Yunhui Guo\textsuperscript{\rm 1}, Yapeng Tian\textsuperscript{\rm 1}\\
\textsuperscript{\rm 1} The University of Texas at Dallas, \textsuperscript{\rm 2} Carnegie Mellon University\\
{\tt\small \{weiguo.pian,yunhui.guo,yapeng.tian\}@utdallas.edu, shentonm@andrew.cmu.edu}
}

\maketitle
\ificcvfinal\thispagestyle{empty}\fi

\renewcommand{\thefootnote}{\fnsymbol{footnote}}
\footnotetext[2]{Equal contribution.}

\begin{abstract}
   In this paper, we introduce audio-visual class-incremental learning, a class-incremental learning scenario for audio-visual video recognition. We demonstrate that joint audio-visual modeling can improve class-incremental learning, but current methods fail to preserve semantic similarity between audio and visual features as incremental step grows. Furthermore, we observe that audio-visual correlations learned in previous tasks can be forgotten as incremental steps progress, leading to poor performance. To overcome these challenges, we propose AV-CIL, which incorporates Dual-Audio-Visual Similarity Constraint (D-AVSC) to maintain both instance-aware and class-aware semantic similarity between audio-visual modalities and Visual Attention Distillation (VAD) to retain previously learned audio-guided visual attentive ability. We create three audio-visual class-incremental datasets, AVE-Class-Incremental (AVE-CI), Kinetics-Sounds-Class-Incremental (K-S-CI), and VGGSound100-Class-Incremental (VS100-CI) based on the AVE, Kinetics-Sounds, and VGGSound datasets, respectively. Our experiments on AVE-CI, K-S-CI, and VS100-CI demonstrate that AV-CIL significantly outperforms existing class-incremental learning methods in audio-visual class-incremental learning. Code and data are available at: \url{https://github.com/weiguoPian/AV-CIL_ICCV2023}.
\end{abstract}


\section{Introduction}
Human perception of the environment is based on a variety of senses. Specifically, people perceive the world by seeing and listening to the events happening around them, which are two of the most commonly used signals for environment perception~\cite{bulkin2006seeing,mo2022multi,mo2023diffava}. By jointly receiving visual and auditory signals, the human brain can better understand its surroundings. Researchers have been inspired by this practical approach to real-world perception and have begun to focus on audio-visual scene understanding. The goal is to guide machines to better perceive their surroundings by learning from audio and visual information jointly, just as humans do.
In recent years, a lot of works have been explored in the field of audio-visual scene understanding, such as audio-visual event localization~\cite{lin2019dual,tian2018audio,tian2019audioevent,wu2019dual}, audio-visual video parsing~\cite{lin2021exploring,mo2022multi,tian2020unified,wu2021exploring}, audio-visual sound separation~\cite{gan2020music,gao2018learning,tian2021cyclic,zhou2020sep}, audio-visual video captioning~\cite{rahman2019watch,tian2019audiocaption,wang2018watch}, and audio-visual sound source localization~\cite{chen2021localizing,chen2022sound,hu2019deep,hu2022mix,li2021space,mo2022localizing,senocak2018learning}. These works have shown that modeling audio-visual modalities jointly can capture cross-modal semantic correlations effectively.
Motivated by the success of these works, we aim to utilize the advantages of audio-visual modeling to mitigate the catastrophic forgetting problem~\cite{li2017learning,mccloskey1989catastrophic,rebuffi2017icarl} in class-incremental learning.

\begin{figure}
    \centering
    \includegraphics[width=0.48\textwidth]{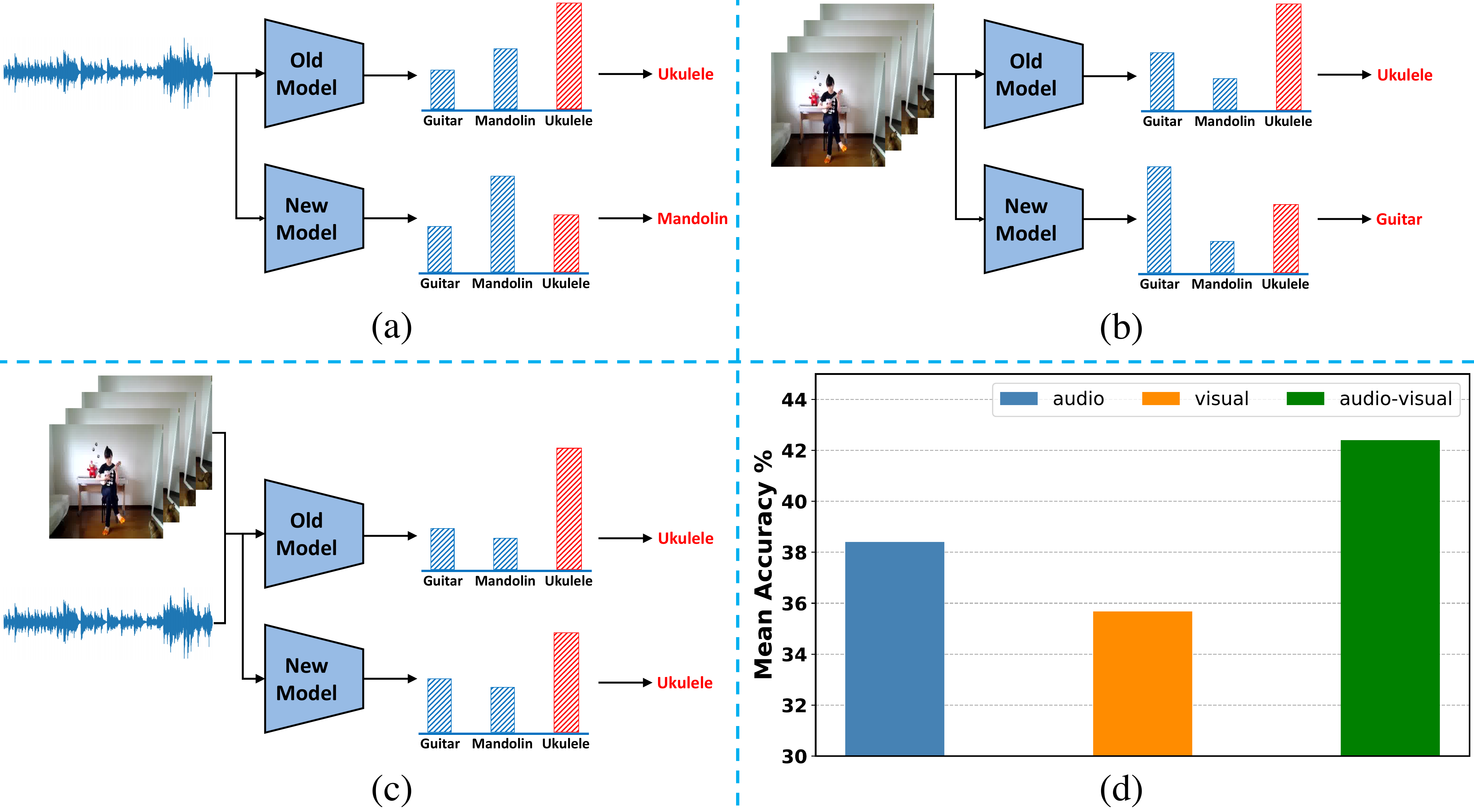}
    \caption{Illustration of training the model in a class-incremental manner using (a) audio modality, (b) visual modality, and (c) joint audio-visual modalities.
    (d) Mean accuracy of using audio modality, visual modality, and joint audio-visual modalities in class-incremental learning with a vanilla fine-tuning strategy on the AVE-CI dataset. Our results show that joint audio-visual modeling can significantly improve perception in the class-incremental setting.}
    \label{fig:advantage_of_av}
    \vspace{-20pt}
\end{figure}

Catastrophic forgetting means the model's performance on previous classes/tasks degrades significantly when updating it by training with only new classes'/tasks' data (or combined with a small set of previous classes’/tasks’ data), which is the key challenge in incremental/continual learning. To alleviate it, a lot of works have been conducted in recent years, which can be categorized into parameter regularization-based~\cite{ahn2019uncertainty,benjamin2019measuring,chaudhry2018riemannian,jung2020continual,kirkpatrick2017overcoming,zenke2017continual}, knowledge distillation-based~\cite{castro2018end,dhar2019learning,douillard2020podnet,hou2019learning,kang2022class,lee2019overcoming,li2017learning,rebuffi2017icarl,wu2019large,zhao2020maintaining,zhu2021class,zhu2021prototype}, replay-memory/exemplar-based~\cite{ahn2021ssil,chaudhry2019efficient,chaudhry2019tiny,kang2022class,lopez2017gradient,rebuffi2017icarl,shin2017continual}, and dynamic architecture-based methods~\cite{golkar2019continual,hung2019compacting,li2019learn,wang2022foster,yan2021dynamically,yoon2018lifelong}, mainly focus on continual/incremental image classification~\cite{ahn2021ssil,douillard2020podnet,rebuffi2017icarl,zhu2021class}, action recognition~\cite{li2021else,park2021class}, semantic segmentation~\cite{cermelli2022incremental,cha2021ssul,douillard2021plop,goswami2023attribution,maracani2021recall,oh2022alife}, object detection~\cite{dong2021bridging,feng2022overcoming,joseph2021incremental}, language/joint-vision-language tasks~\cite{ke2023continual,mi2020continual,scialom2022fine,srinivasan2022climb}, and self-supervised representation learning/pre-training~\cite{fini2022self,jang2022towards,madaan2022representational,purushwalkam2022challenges,yan2022generative}. 
Despite the success of audio-visual modeling in capturing cross-modal semantic correlations, its potential in addressing the catastrophic forgetting problem in class-incremental learning remains unexplored.


In order to explore the effectiveness of joint audio-visual modeling in class-incremental learning, we first train the model using a vanilla fine-tuning strategy, which involves fine-tuning the model on new classes directly without any specific techniques (please see Section~\ref{sec:experiments} for experimental settings). The results are presented in Figure~\ref{fig:advantage_of_av}, which demonstrates the advantage of joint audio-visual modeling over single audio or visual modality in a class-incremental manner. Based on these findings, we propose Audio-Visual Class-Incremental Learning, which is a novel incremental learning problem under the scenario of audio-visual video recognition.
Since both audio and visual modalities are involved, simply applying existing class-incremental learning approaches to our new audio-visual task cannot fully exploit the natural cross-modal association between the two modalities. Instead, we propose that the correlation preservation between visual and audio modalities should be explicitly incorporated to further improve the current techniques for incremental learning of audio-visual data.

In this paper, we propose a method named AV-CIL (\textbf{A}udio-\textbf{V}isual\textbf{-}\textbf{C}lass-\textbf{I}ncremental \textbf{L}earning) to address our new audio-visual class-incremental learning problem. In AV-CIL, we introduce a \textit{Dual-Audio-Visual Similarity Constraint (D-AVSC)}, which is designed to preserve both instance-aware and class-aware semantic similarities between audio and visual modalities throughout the increase in the number of classes.

Moreover, to better learn the cross-modal feature correlations, and ultimately get better joint audio-visual features, we also adopt an audio-guided visual attention mechanism~\cite{li2021space,tian2018audio,tian2021can,wu2019dual}, which is an effective attention mechanism for adaptively learning correlations between audio and visual features. However, under the incremental learning settings, with the increasing of the incremental steps, we observe that the learned audio-visual attentive ability in previous tasks could get vanished, which results in the forgetting of learned audio-visual correlations in previous tasks. Please see Figure~\ref{fig:attn_distill_motivation} for the visualization of this phenomenon. To preserve and leverage the previously learned attentive ability in future classes/tasks, we propose the \textit{Visual Attention Distillation (VAD)} to distil the learned audio-guided visual attentive ability into new incremental steps, which enables the model to preserve previously learned cross-modal audio-visual correlations in new classes/tasks.

We use three existing audio-visual datasets: AVE~\cite{tian2018audio}, Kinetics-Sounds~\cite{arandjelovic2017look}, and VGGSound~\cite{chen2020vggsound} to construct datasets for class-incremental learning. We name the three newly constructed datasets as 
AVE-Class-Incremental (AVE-CI), Kinetics-Sounds-Class-Incremental (K-S-CI), and VGGSound100-Class-Incremental (VS100-CI), respectively. We conduct experiments on the three datasets to evaluate the effectiveness of our method in audio-visual class-incremental learning. The experimental results show that our proposed method outperforms state-of-the-art class-incremental learning methods significantly on all three datasets.
In summary, this paper contributes follows: 
\begin{itemize}
    \item To explore the effectiveness of joint audio-visual modeling in the alleviation of the catastrophic forgetting problem in class-incremental learning, we propose audio-visual class-incremental learning that trains the model continually under the scenario of audio-visual video recognition.  To the best of our knowledge, this is the first work on audio-visual incremental learning.  
    \item We propose a method, named \textit{AV-CIL}, to tackle the posed new problem. AV-CIL contains a \textit{Dual-Audio-Visual Similarity Constraint (D-AVSC)} to preserve both instance- and class-aware semantic similarity between audio and visual features throughout the incremental steps. 
    Furthermore, we also propose \textit{Visual Attention Distillation (VAD)} to enable the model to preserve 
    previously learned attentive ability in future classes/tasks
    for preventing the model from forgetting previously learned audio-visual correlations.
    \item Experimental results on three audio-visual class-incremental datasets, AVE-CI, K-S-CI, and VS100-CI (constructed from AVE~\cite{tian2018audio}, Kinetics-Sounds~\cite{arandjelovic2017look}, and VGGSound~\cite{chen2020vggsound}), demonstrate that our method outperforms state-of-the-art class-incremental learning methods significantly.
\end{itemize}

\section{Related Work}
\subsection{Audio-Visual Learning}
Audio-visual data can provide more synchronized and/or complementary information than unimodal audio or visual data, which has been proven to be more effective in scene understanding tasks, such as visual-guided sound separation/localization~\cite{chen2021localizing,chen2022sound,gan2020music,gao2018learning,hu2019deep,hu2022mix,li2021space,mo2022SLAVC,mo2022localizing,mo2023unified,mo2023audiovisual,mo2023avsam,senocak2018learning,tian2021cyclic,zhou2020sep}, audio-visual video parsing~\cite{lin2021exploring,mo2022multi,tian2020unified,wu2021exploring},  audio-visual event localization~\cite{lin2019dual,tian2018audio, tian2019audioevent, wu2019dual}, audio-visual video caption~\cite{rahman2019watch,tian2019audiocaption,wang2018watch}, audio-visual navigation~\cite{chen2021semantic,chen2020soundspaces,chen2021learning}, etc. 
Lately, researchers have also been paying attention to generalizable audio-visual representation learning methods, \textit{e.g.} audio-visual zero-shot learning methods~\cite{mazumder2021avgzslnet,mercea2022temporal}, and joint audio-visual pre-training~\cite{gong2023contrastive}. A recent survey on audio-visual learning can be found in \cite{wei2022learning}.
In this work, we mainly focus on audio-visual learning under class-incremental settings, in which the model continually learns data from new tasks/classes, to explore the feasibility of using cross-modal audio-visual correlations to mitigate the catastrophic forgetting problem in class-incremental learning.

\subsection{Incremental Learning}
\noindent
\textbf{Parameter Regularization.} Parameter regularization-based methods~\cite{ahn2019uncertainty,benjamin2019measuring,chaudhry2018riemannian,jung2020continual,kirkpatrick2017overcoming,zenke2017continual} aims to estimate the importance of different parameters of the model and allocate them with different weights to indicate their importance. During incremental steps, unimportant parameters can be updated much easier than important parameters.

\noindent
\textbf{Knowledge Distillation.}
Knowledge distillation-based methods~\cite{castro2018end,dhar2019learning,douillard2020podnet,hou2019learning,kang2022class,lee2019overcoming,li2017learning,rebuffi2017icarl,wu2019large,zhao2020maintaining,zhu2021class,zhu2021prototype} help the model preserve previously learned knowledge in current/future incremental steps, which can be implemented as the minimization of the distance between the representations generated by the previous and current model~\cite{douillard2020podnet,kang2022class}, or by minimizing the divergence (\textit{e.g.} Kullback-Leibler divergence) between the output probability distribution of the previous and current model~\cite{ahn2021ssil,li2017learning,rebuffi2017icarl}.

\noindent
\textbf{Exemplar/Memory Replay.}
Replay-based methods assume that small size of memory is accessible to store examples from old tasks/classes~\cite{ahn2021ssil,chaudhry2019efficient,chaudhry2019tiny,kang2022class,lopez2017gradient,rebuffi2017icarl,shin2017continual}. iCaRL~\cite{rebuffi2017icarl} first proposed the nearest-mean-of-exemplars selection strategy to select the most representative exemplars in each class, which is also followed by later works~\cite{kang2022class,park2021class}. On the other hand, pseudo-rehearsal techniques~\cite{odena2017conditional,ostapenko2019learning} were also proposed to use generative models to generate pseudo-exemplars based on the estimated distribution of data from previous classes.

\noindent
\textbf{Dynamic Architecture.}
Architecture-based methods~\cite{golkar2019continual,hung2019compacting,li2019learn,wang2022foster,yan2021dynamically,yoon2018lifelong} hold incremental modules to increase the capacity of the model to handle new tasks/classes. Conventional architecture-based methods may cause unaffordable overhead caused by adding new modules continually as the incremental step grows~\cite{yan2021dynamically}. Recently, the combination between dynamic architecture and distillation alleviates the continual-increasing overhead problem by distilling the increased modules into the original volume~\cite{wang2022foster}.

\noindent
\textbf{Existing Fields of Incremental Learning.}
The aforementioned incremental learning methods or techniques mainly focus on the fields of image classification~\cite{ahn2021ssil,douillard2020podnet,guo2020improved,rebuffi2017icarl,yu2023scale,zhu2021class}, action recognition~\cite{li2021else,park2021class}, semantic segmentation~\cite{cermelli2022incremental,cha2021ssul,douillard2021plop,goswami2023attribution,maracani2021recall,oh2022alife}, object detection~\cite{dong2021bridging,feng2022overcoming,joseph2021incremental}, language/joint-vision-language tasks~\cite{ke2023continual,mi2020continual,scialom2022fine,srinivasan2022climb}, and self-supervised representation learning/pre-training~\cite{fini2022self,jang2022towards,madaan2022representational,purushwalkam2022challenges,yan2022generative}. 
In this paper, we focus on audio-visual class-incremental learning, in which we try to use the strengths of joint audio-visual modeling to alleviate the catastrophic forgetting problem in class-incremental learning.





\section{Method}

\begin{figure*}
    \centering
    \includegraphics[width=0.98\textwidth]{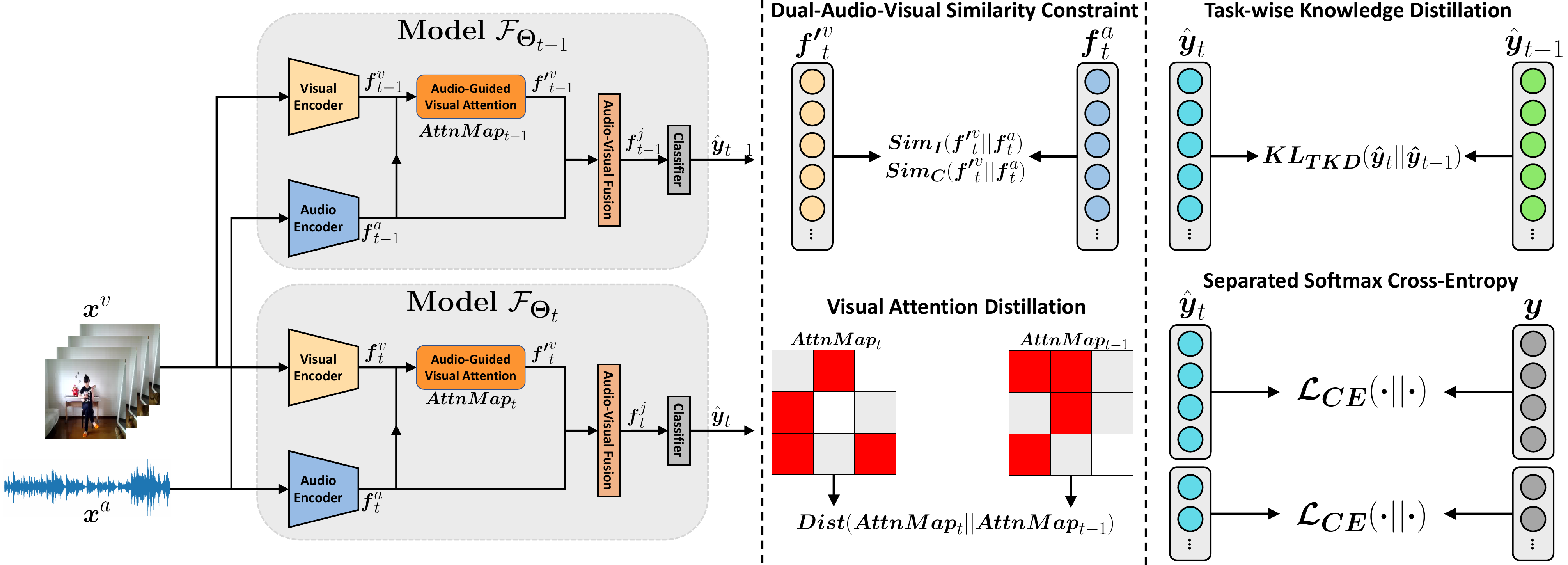}
    \caption{Overview of our proposed AV-CIL, which consists of four main components: Dual-Audio-Visual Similarity Constraint (D-AVSC), Visual Attention Distillation (VAD), Task-wise Knowledge Distillation~\cite{ahn2021ssil,li2017learning}, and Separated Softmax Cross-Entropy~\cite{ahn2021ssil}.}
    \label{fig:overview}
    \vspace{-10pt}
\end{figure*}

\subsection{Problem Formulation}
Class-incremental learning aims to train the model $\mathcal{F}_{\boldsymbol{\Theta}}$ with parameters $\boldsymbol{\Theta}$ through a sequence of $T$ tasks $\{\mathcal{T}_1,\mathcal{T}_2,...,\mathcal{T}_T\}$. In our audio-visual class-incremental learning, for a task $\mathcal{T}_t$ (incremental step $t$), its corresponding training set can be denoted as $\mathcal{D}_t=\{(\boldsymbol{x}^a_{t,i}, \boldsymbol{x}^v_{t,i}, y_{t,i})\}_{i=1}^{n_t}$, where $\boldsymbol{x}^a_{t,i}$ and $\boldsymbol{x}^v_{t,i}$ denote the $i^{th}$ input sample's audio and visual modalities respectively in $\mathcal{D}_t$, and $y_{t,i}\in\mathcal{C}_t$ is the corresponding label, where $\mathcal{C}_t$ denotes the label space of task $\mathcal{T}_t$.
For any two tasks' training label space $\mathcal{C}_{t_1}$ and $\mathcal{C}_{t_2}$, we have $\mathcal{C}_{t_1}\cap\mathcal{C}_{t_2}=\emptyset$. In our settings, storing a small fixed size of exemplar/memory set of data from previous classes is permitted, which is denoted as $\mathcal{M}_{t}$. Therefore, all accessible training data at incremental step $t$ can be denoted as $\mathcal{D}'_t=\mathcal{D}_t\cup\mathcal{M}_{t}$. 
Note that, $|\mathcal{M}_1|=0$, and $|\mathcal{M}_t|$ is a fixed number when $t>1$.
We use $|\mathcal{C}_t|$ and $|\mathcal{C}_{\mathcal{M}_t}|$ to denote the number of classes in task $\mathcal{T}_t$ and the number of classes in task $\mathcal{T}_t$'s exemplar set (the number of classes in task $\mathcal{T}_t$'s all previous tasks) respectively.
Therefore, the exemplar number of each class in $\mathcal{M}_t$ can be presented as $m_t=|\mathcal{M}_t|/|\mathcal{C}_{\mathcal{M}_t}|$.
We also let $|\mathcal{C}'_{t}|$ be the number of all classes up to the incremental step $t$, \textit{i.e.} $|\mathcal{C}'_t|=|\mathcal{C}_t|+|\mathcal{C}_{\mathcal{M}_t}|$. Thus, the training process at incremental step $t$ can be formulated as:
\begin{equation}
    \begin{split}
        \boldsymbol{\Theta}_{t}=\argmin_{\boldsymbol{\Theta}_{t-1}} \mathbb{E}_{(\boldsymbol{x}^a,\boldsymbol{x}^v,y)\sim\mathcal{D}'_t}[\mathcal{L}(\mathcal{F}_{\boldsymbol{\Theta}_{t-1}}(\boldsymbol{x}^a,\boldsymbol{x}^v),y)],
    \end{split}
    \label{eq:absolute_pos}
\end{equation}
where $\mathcal{L}$ is the loss function between the model's outputs and the corresponding labels. After the training process on $\mathcal{D}_t'$, the model is evaluated on the testing set that contains data from all seen classes up to step $t$.

\subsection{Overview}
The overview of our proposed method, AV-CIL, is illustrated in Figure~\ref{fig:overview}. Our method mainly consists of Task-wise Knowledge Distillation~\cite{ahn2021ssil,li2017learning}, Separated Softmax Cross-Entropy~\cite{ahn2021ssil}, and our proposed Dual-Audio-Visual Similarity Constraint (D-AVSC) and Visual Attention Distillation (VAD). The model used in our method is composed of a visual encoder, an audio encoder, an audio-guided visual attention layer~\cite{li2021space,tian2018audio,tian2021can,wu2019dual}, an audio-visual fusion layer, and a classifier. For the audio encoder and the visual encoder, inspired by the excellent performance and generalization capability of recent vision/audition self-supervised pre-trained models~\cite{chen2020simple,he2022masked,he2020momentum,huang2022masked,tong2022videomae}, we apply the self-supervised pre-trained AudioMAE~\cite{huang2022masked} and VideoMAE~\cite{tong2022videomae} as the audio encoder and the visual encoder, respectively. Moreover, due to their label- and class-free pre-training scheme, the self-supervised pre-trained models are suitable for incremental learning settings. 

Given an input sequence of frames $\boldsymbol{x}^v$ and the associated audio signal $\boldsymbol{x}^a$, we first use the audio encoder (AudioMAE) and the visual encoder (VideoMAE) to generate the audio feature $\boldsymbol{f}^a\in\mathbb{R}^{d}$ and the visual feature $\boldsymbol{f}^v\in\mathbb{R}^{L\times S\times d}$ respectively, where $L$ and $S$ denote the temporal and spatial dimension of the visual feature respectively. After that, the audio-guided visual attention layer is used to pool the visual feature by both spatial and temporal dimensions, which is an effective mechanism to \textit{adaptively} learn correlations between audio and visual features~\cite{li2021space,tian2018audio,tian2021can,wu2019dual}.

Firstly, we calculate the spatial attention map on each temporal frame through the following process:
\begin{equation}
    \begin{split}
        \boldsymbol{Score}^a =& \sigma(\boldsymbol{f}^a\boldsymbol{W}^a), \\
        \boldsymbol{Score}^v_l =& \sigma(\boldsymbol{f}^v_l\boldsymbol{W}^v), \\
        \boldsymbol{w}^{Spa.}_l = Softmax&(\boldsymbol{Score}^a\odot\boldsymbol{Score}^v_l),
    \end{split}
    \label{eq:spa_attention}
\end{equation}
where $\boldsymbol{W}^a\in\mathbb{R}^{d\times d}$ and $\boldsymbol{W}^v\in\mathbb{R}^{d\times d}$ are the learnable projection matrices for audio and visual features respectively, and $\sigma(\cdot)$ is the nonlinear activation unit, \textit{e.g.} Tanh function. $\boldsymbol{f}^v_l\in\mathbb{R}^{S\times d}$ and $\boldsymbol{w}^{Spa.}_l\in\mathbb{R}^{S\times d}$ denote the $l^{th}$ temporal frame's feature map and spatial attention map respectively, $\odot$ denotes the Hadamard product.
Based on the calculated spatial attention map of each temporal frame, the temporal attention map calculation can be formulated as:
\begin{equation}
    \begin{split}
        \boldsymbol{Score'}^v_l = \sum&(\boldsymbol{w}^{Spa.}_l\odot\boldsymbol{Score}^v_l), \\
        \boldsymbol{w}^{Tem.}= Softmax&([\boldsymbol{Score'}^v_1,...,\boldsymbol{Score'}^v_L]),
    \end{split}
    \label{eq:tem_attention}
\end{equation}
where $\boldsymbol{Score'}^v_l \in \mathbb{R}^d$. $\boldsymbol{w}^{Tem.}\in\mathbb{R}^{L\times d}$ denotes the temporal attention map. With the calculated spatial and temporal attention map, we can pool the original visual feature to get the attend visual feature map with more audio-visual cross-modal correlations:
\begin{equation}
    \begin{split}
        \boldsymbol{f'}^v=\sum^L_{l=1}\boldsymbol{w}^{Tem.}_l\odot \sum(\boldsymbol{f}^v_l\odot\boldsymbol{w}^{Spa.}_l),
    \end{split}
    \label{eq:visual_pooling}
\end{equation}
After the above process, an audio-visual fusion layer follows to get the joint audio-visual features that will be fed into the classifier to get the final results:
\begin{equation}
    \begin{split}
        \hat{y}=CLS(\sigma(\boldsymbol{f}^a\boldsymbol{U}^a) + \sigma(\boldsymbol{f'}^v\boldsymbol{U}^v)),
    \end{split}
    \label{eq:fusion_classify}
\end{equation}
where $\boldsymbol{U}^a\in\mathbb{R}^{d\times d}$ and $\boldsymbol{U}^v\in\mathbb{R}^{d\times d}$ are the projection matrices, and $CLS$ denotes the classifier.

\subsection{Dual-Audio-Visual Similarity Constraint}
In this subsection, we will introduce our proposed Dual-Audio-Visual Similarity Constraint (D-AVSC), which aims to preserve the cross-modal semantic similarity, which is crucial to audio-visual modeling~\cite{mo2022multi,tian2021can}, throughout the incremental steps.

Our D-AVSC is a dual constraint consisting of an Instance-aware Audio-Visual Semantic Similarity (I-AVSS) and a Class-aware Audio-Visual Semantic Similarity (C-AVSS). In I-AVSS, our goal is to maximize the similarity between audio and visual semantic features extracted from the \textit{same video sample}, while minimizing the cross-modal similarity between audio and visual semantic features obtained from \textit{different video samples}.
In the incremental step $t$, where $t > 1$, our I-AVSS can be formulated as:
\begin{equation}
    \begin{split}
        \mathcal{L}_{I}=-\mathbb{E}_{(\boldsymbol{x}_i^a,\boldsymbol{x}_i^v)\sim\mathcal{D}'_t}\left[\text{log}\frac{e^{\boldsymbol{f}^a_{t,i} {\boldsymbol{f'}^v_{t,i}}^\text{T}/\tau}}{\sum_{j=1}^N e^{\boldsymbol{f}^a_{t,i} {\boldsymbol{f'}^v_{t,j}}^\text{T}/\tau}}\right],
    \end{split}
    \label{eq:I-AVSS}
\end{equation}
where $N$ denotes the number of samples in a mini-batch. Our C-AVSS aims to preserve the cross-modal semantic similarities/dissimilarities among samples from the same/different classes. Specifically, it can maximize the similarity between audio and visual semantic features belonging to the \textit{same class} and minimize the similarity between audio and visual features from \textit{different classes}. Our C-AVSS can be expressed as follows during the incremental step $t$:
\begin{equation}
    \begin{split}
        \mathcal{L}_{C}=&-\mathbb{E}_{(\boldsymbol{x}_i^a,\boldsymbol{x}_i^v,y_i)\sim\mathcal{D}'_t} \\
        &\left[\text{log}\frac{\sum_{j=1}^N e^{\boldsymbol{f}^a_{t,i} {\boldsymbol{f'}^v_{t,j}}^\text{T}/\tau}\cdot\mathbbm{1}[y_i=y_j]}{(\sum^N_{j=1}\mathbbm{1}[y_i=y_j]) (\sum_{j=1}^N e^{\boldsymbol{f}^a_{t,i} {\boldsymbol{f'}^v_{t,j}}^\text{T}/\tau})}\right],
    \end{split}
    \label{eq:C-AVSS}
\end{equation}
where $\mathbbm{1}[y_i=y_j]$ is an indicator that equals to 1 iff $y_i=y_j$. Our Dual-Audio-Visual Similarity Constraint (D-AVSC) is formulated as:
\begin{equation}
    \begin{split}
        \mathcal{L}_{D-AVSC}=\lambda_I\mathcal{L}_I + \lambda_C\mathcal{L}_C,
    \end{split}
    \label{eq:D-AVSC}
\end{equation}
where $\lambda_I$ and $\lambda_C$ are hyperparameters to balance the two constraint loss values.

\begin{figure*}
    \centering
    \includegraphics[width=0.98\textwidth]{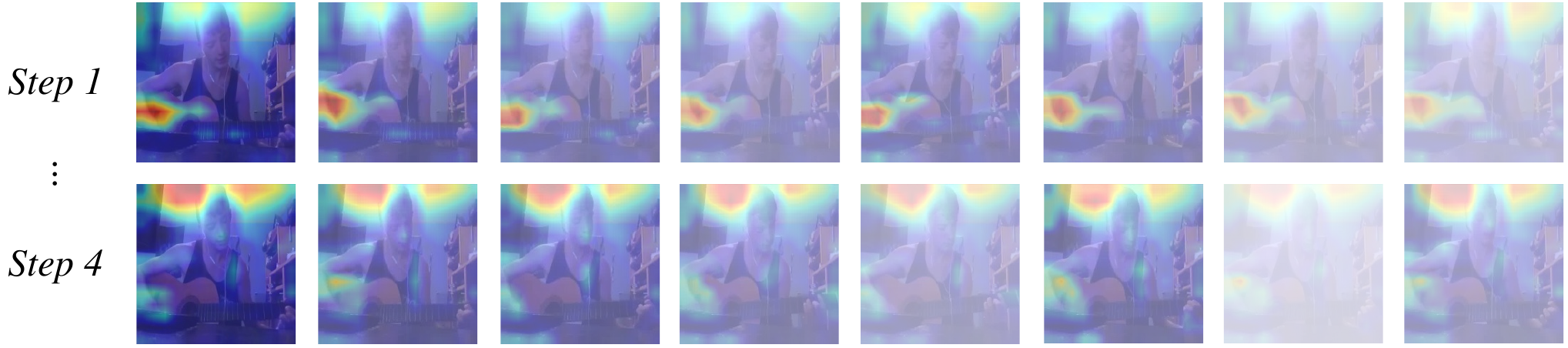}
    \caption{Visualization of the vanishing of visual attention as the incremental step increases. With the classes increasing, previously learned audio-guided visual attentive abilities, as well as the learned audio-visual correlations, are getting forgotten.}
    \label{fig:attn_distill_motivation}
\end{figure*}

\subsection{Visual Attention Distillation}
As we mentioned before, the learned audio-visual correlation could vanish as the incremental step increases. We visualize this phenomenon in Figure~\ref{fig:attn_distill_motivation}, which is generated by the model trained without our proposed Visual Attention Distillation process. The example is randomly selected from the validation set. In the figure, the heatmap in each frame shows the attentive level of different spatial regions by the corresponding audio, and the transparency of each frame demonstrates its temporal attentive level. 
We can see a man playing guitar in the video. Before classes increase, the most attentive region in each frame is around the area where the hand strums strings. However, as the incremental step grows, the most attentive region in each frame changes, as well as the temporal attentive level of each temporal frame. 
To address this problem, we propose the \textit{Visual Attention Distillation (VAD)} to enable the model to preserve and leverage the previously learned attentive ability in future classes/tasks. In this way, the model can still maintain the visual attention ability without forgetting. For the full version of the visualization, please see Appendix for details.

Specifically, during an incremental step $t$ where $t>1$, we use the spatial and temporal visual attention maps generated by the model learned from the last incremental step $t-1$ as the target for the knowledge distillation process of current learned attention maps. Note that we only distill the visual attention map for data from the exemplar set. Our VAD is composed of the spatial distillation part and the temporal distillation part. For the spatial distillation part, we use the following formulation to express it:
\begin{equation}
    \begin{split}
        Dist_{spa.}=\mathbb{E}_{(\boldsymbol{x}^a,\boldsymbol{x}^v)\sim\mathcal{M}_t} \Big[KL(\boldsymbol{w}^{Spa.}_{t}||\boldsymbol{w}^{Spa.}_{t-1}) \Big],
    \end{split}
    \label{eq:dist_spa}
\end{equation}
where $\boldsymbol{w}_t^{Spa.}=[\boldsymbol{w}_{t,1}^{Spa.},\boldsymbol{w}_{t,2}^{Spa.},...,\boldsymbol{w}_{t,L}^{Spa.}]$ and $\boldsymbol{w}_{t-1}^{Spa.}=[\boldsymbol{w}_{t-1,1}^{Spa.},\boldsymbol{w}_{t-1,2}^{Spa.},...,\boldsymbol{w}_{t-1,L}^{Spa.}]$ denote the spatial attention maps for all $L$ temporal frames generated by model $\mathcal{F}_{\boldsymbol{\Theta}_t}$ and model $\mathcal{F}_{\boldsymbol{\Theta}_{t-1}}$ respectively. $KL(\cdot||\cdot)$ denotes the Kullback-Leibler (KL) divergence function. Similarly, the temporal distillation part can be formulated as:
\begin{equation}
    \begin{split}
        Dist_{tem.}=\mathbb{E}_{(\boldsymbol{x}^a,\boldsymbol{x}^v)\sim\mathcal{M}_t} \Big[KL(\boldsymbol{w}^{Tem.}_{t}||\boldsymbol{w}^{Tem.}_{t-1}) \Big],
    \end{split}
    \label{eq:dist_tem}
\end{equation}
where $\boldsymbol{w}^{Tem.}_t$ and $\boldsymbol{w}^{Tem.}_{t-1}$ are the temporal attention maps generated by model $\mathcal{F}_{\boldsymbol{\Theta}_t}$ and model $\mathcal{F}_{\boldsymbol{\Theta}_{t-1}}$ respectively. Finally, the overall of our VAD can be illustrated by combining the above two parts:
\begin{equation}
    \begin{split}
        \mathcal{L}_{VAD}=\lambda_{VAD}Dist_{spa.} + (1-\lambda_{VAD})Dist_{tem.}.
    \end{split}
    \label{eq:VAD}
\end{equation}

\subsection{Final Loss Function}
Above, we introduce our proposed Dual-Audio-Visual Similarity Constraint (D-AVSC) and Visual Attention Distillation (VAD). For our final optimization objective, we combine our D-AVSC and VAD with the Separated Softmax Cross-Entropy (SS-CE)~\cite{ahn2021ssil} and the Task-wise Knowledge Distillation (TKD)~\cite{ahn2021ssil,castro2018end,li2017learning}. 

SS-CE~\cite{ahn2021ssil} is a modified version of the original softmax loss function for class-incremental learning, aims to prevent the prediction score of old class from being penalized by training on new tasks, which can be denoted as:
\begin{equation}
    \begin{split}
        \mathcal{L}_{SS-CE}=\mathbb{E}_{(\boldsymbol{x}^a,\boldsymbol{x}^v,y)\sim\mathcal{D}'_t}& \Big[\mathcal{L}_{CE}(\hat{\boldsymbol{y}}^{\mathcal{C}_t}, \boldsymbol{y}^{\mathcal{C}_t})\cdot\mathbbm{1}[y\in\mathcal{C}_t] \\
        +& \mathcal{L}_{CE}(\hat{\boldsymbol{y}}^{\mathcal{C}_{\mathcal{M}_t}},\boldsymbol{y}^{\mathcal{C}_{\mathcal{M}_t}})\cdot\mathbbm{1}[y\notin\mathcal{C}_t] \Big],
    \end{split}
    \label{eq:SS}
\end{equation}
where $\hat{\boldsymbol{y}}^{\mathcal{C}_t}=\hat{\boldsymbol{y}}_{[|\mathcal{C}_{\mathcal{M}_t}|+1:|\mathcal{C}'_t|]}$, $\hat{\boldsymbol{y}}^{\mathcal{C}_{\mathcal{M}_t}}=\hat{\boldsymbol{y}}_{[1:|\mathcal{C}_{\mathcal{M}_t}|]}$. $\boldsymbol{y}^{\mathcal{C}_t}$ and $\boldsymbol{y}^{\mathcal{C}_{\mathcal{M}_t}}$
are the one-hot vector of $y$ in $\mathbb{R}^{|\mathcal{C}_{\mathcal{M}_t}|+1:|\mathcal{C}'_t|}$ and $\mathbb{R}^{|\mathcal{C}_{\mathcal{M}_t}|}$ respectively.

The last part of our final optimization objective is the Task-wise Knowledge Distillation (TKD)~\cite{ahn2021ssil,castro2018end,li2017learning}, which is used for preserving the task-wise knowledge and preventing the learned knowledge from being biased by other tasks:
\begin{equation}
    \begin{split}
        \mathcal{L}_{TKD}=\mathbb{E}_{(\boldsymbol{x}^a,\boldsymbol{x}^v)\sim\mathcal{D}'_t} \Bigg[\sum_{s=1}^{t} KL(\hat{\boldsymbol{y}}^{\mathcal{C}_s}_t||\hat{\boldsymbol{y}}^{\mathcal{C}_s}_{t-1}) \Bigg],
    \end{split}
    \label{eq:TKD}
\end{equation}
where $\hat{\boldsymbol{y}}_t^{\mathcal{C}_s}$ denotes scores of classes in $\mathcal{C}_s$ of $\hat{\boldsymbol{y}}_t$. Finally, our overall loss function is denoted as:
\begin{equation}
    \begin{split}
        \mathcal{L}_{AV-CIL}=\mathcal{L}_{SS-CE} + \mathcal{L}_{TKD} + \mathcal{L}_{D-AVSC} + \mathcal{L}_{VAD}.
    \end{split}
    \label{eq:overall}
\end{equation}

\subsection{Management of Exemplar Set}
In our proposed method, we hold the memory with a fixed maximum number of exemplars. In each task, we follow~\cite{ahn2021ssil} and randomly select exemplars for new classes, and randomly reduce the number of exemplars in old classes to keep the same exemplars number in all classes. 



\section{Experiments}
\label{sec:experiments}
In this section, we first introduce our experimental setup, including datasets, baseline methods, and hyperparameters. Then, we show the experimental results of our proposed methods compared to the state-of-the-art methods.

\noindent
\textbf{Datasets.} We conduct experiments with our AV-CIL compared to state-of-the-art baselines on the AVE~\cite{tian2018audio}, Kinetics-Sounds~\cite{arandjelovic2017look}, and VGGSound~\cite{chen2020vggsound} datasets. The AVE dataset consists of 4K 10-seconds videos from 28 audio-visual event classes. In our class-incremental settings, we randomly split the 28 audio-visual event classes into 4 incremental tasks, each of which contains 7 classes. We name it as \textit{AVE-Class-Incremental (AVE-CI)}. 
Kinetics-Sounds (K-S), which contains around 24K 10-seconds videos (20K, 2K and 2K for training, validation and testing respectively) from 31 human action classes, is a subset selected from Kinetics-400~\cite{kay2017kinetics} dataset. In our settings, we randomly select 30 classes from the K-S and randomly divide them into 5 incremental steps, each of which contains 6 classes. We name our new constructed dataset as \textit{Kinetics-Sounds-Class-Incremental (K-S-CI)}, which contains around 23K samples in total. 
The VGGSound~\cite{chen2020vggsound} dataset contains around 200K 10-seconds videos from 309 classes. We randomly select 100 classes from the original VGGSound dataset to construct a subset named VGGSound100, which contains 60K samples in total. For each class of the VGGSound100, We randomly select 50 samples for validation and 50 samples for testing.
In our class-incremental settings, we randomly divide the 100 classes into 10 incremental steps, each of which contains 10 classes. We name it as \textit{VGGSound100-Class-Incremental (VS100-CI)}.

\noindent
\textbf{Baselines.} We compare our proposed method with following representative and state-of-the-art methods: Fine-tuning, LwF~\cite{li2017learning}, iCaRL~\cite{rebuffi2017icarl}, SS-IL~\cite{ahn2021ssil}, and AFC~\cite{kang2022class}. Fine-tuning is the simplest incremental training strategy that initialized the model with the parameters trained from the last task and re-train it on the current task without any constraints or other strategies to prevent the model from catastrophic forgetting.
For iCaRL~\cite{rebuffi2017icarl}, we report the experimental results with both nearest-mean-of-exemplars (NME)~\cite{rebuffi2017icarl} classification strategy and the classifier. We name them as iCaRL-NME and iCaRL-FC, respectively. Note that, iCaRL-FC is also named as iCaRL-CNN in other papers~\cite{kang2022class}. For AFC~\cite{kang2022class}, we also report the experimental results with both NME classification strategy and the classifier, which are named AFC-NME and AFC-LSC respectively in our paper. We also present the experimental results of the Oracle/Upper Bound, which is defined as using all training data from seen classes to train the model. Please note that, for fair comparisons, all baselines use the same backbone as our method, including the visual encoder, audio encoder, audio-guided visual attention layer, and audio-visual fusion layer. Moreover, for exemplar-based methods iCaRL~\cite{rebuffi2017icarl}, SS-IL~\cite{ahn2021ssil}, and AFC~\cite{kang2022class}, the exemplar selection strategies keep same as per their original papers, and all of them use the same memory size as ours.


\begin{table*}[htbp]
  \centering
  \caption{Audio-visual class-incremental results of different approaches on the AVE-CI, K-S-CI, and VS100-CI datasets. The bold part denotes the overall best results, and the underlined part denotes the best results of baselines. Our AV-CIL achieves the best performance on all three datasets.}
    \begin{tabular}{lcp{7.25em}cp{7.75em}cc}
    \toprule
    \multirow{2}[4]{*}{Methods} & \multicolumn{2}{c}{AVE-CI} & \multicolumn{2}{c}{K-S-CI} & \multicolumn{2}{c}{VS100-CI} \\ \cmidrule(r){2-3} \cmidrule(r){4-5} \cmidrule(r){6-7}   & Mean Acc. & \multicolumn{1}{c}{Ave. Forget.} & Mean Acc. & \multicolumn{1}{c}{Ave. Forget.} & Mean Acc. & Ave. Forget. \\
    \midrule
    Fine-tuning & 42.40  & \multicolumn{1}{c}{70.99} & 41.18 & \multicolumn{1}{c}{89.62} & 26.21 & 89.37 \\
    LwF   & 58.07 & \multicolumn{1}{c}{26.90} & 65.54 & \multicolumn{1}{c}{16.55} & 59.34 & 23.01 \\
    iCaRL-NME & 56.15 & \multicolumn{1}{c}{\underline{11.71}} & 64.51 & \multicolumn{1}{c}{18.70} & 56.19 & 12.80 \\
    iCaRL-FC & 65.88 & \multicolumn{1}{c}{26.08} & 65.54 & \multicolumn{1}{c}{40.57} & 64.22 & 29.94 \\
    SS-IL & 61.94 & \multicolumn{1}{c}{22.49} & \underline{69.71} & \multicolumn{1}{c}{\underline{10.53}} & \underline{69.20}  & \underline{9.75} \\
    AFC-NME & \underline{68.46} & \multicolumn{1}{c}{14.18} & 69.13 & \multicolumn{1}{c}{27.30} & 61.41 & 23.30 \\
    AFC-LSC & 65.21 & \multicolumn{1}{c}{28.11} & 67.02 & \multicolumn{1}{c}{33.56} & 57.76 & 29.64 \\
    AV-CIL (Ours) & \textbf{74.04} & \multicolumn{1}{c}{\textbf{7.63}} & \textbf{73.06} & \multicolumn{1}{c}{\textbf{6.48}} & \textbf{72.80}  & \textbf{5.49} \\
    \midrule
    Oracle (Upper Bound) & 76.85 & \multicolumn{1}{c}{--}  & 80.43 & \multicolumn{1}{c}{--}  & 78.63 & \multicolumn{1}{c}{--} \\
    \bottomrule
    \end{tabular}%
  \label{tab:main_res}%
\end{table*}%

\begin{figure*}[htbp]
    \centering
    \subfloat[]{
    \centering
    \includegraphics[width=0.325\textwidth]{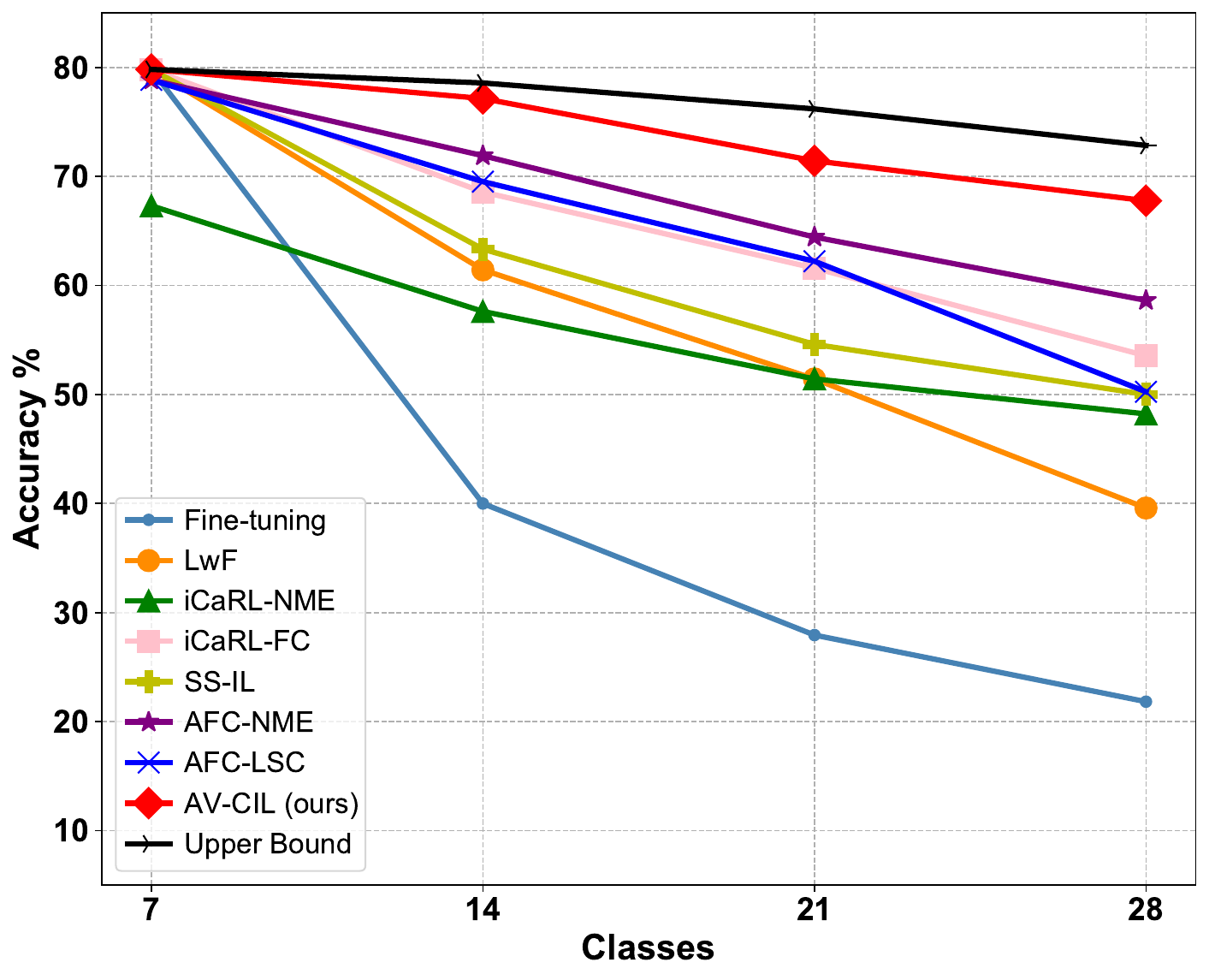}
    \label{fig:acc_AVE_av}
    }
    \subfloat[]{
    \centering
    \includegraphics[width=0.325\textwidth]{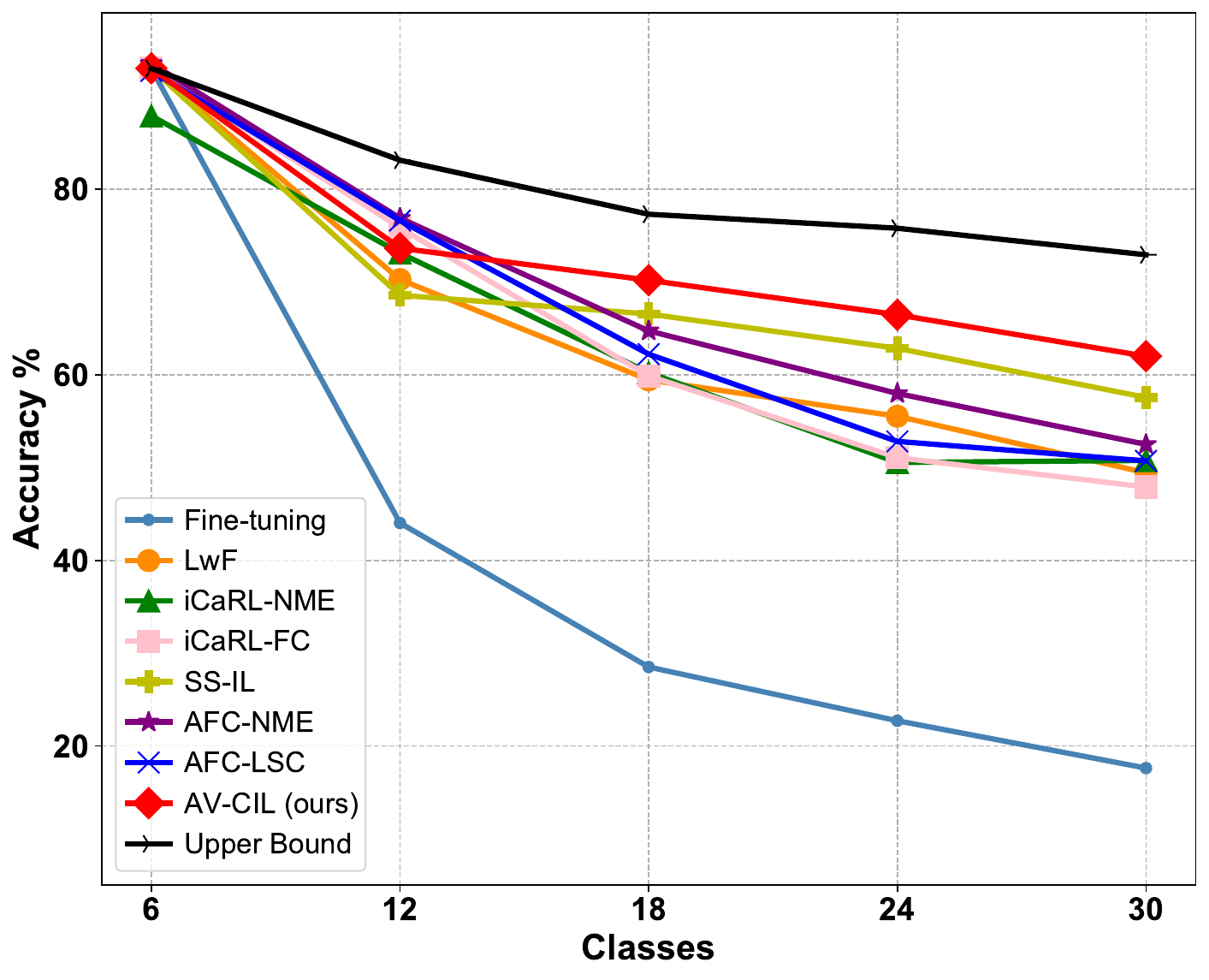}
    \label{fig:acc_KS_av}
    }
    \subfloat[]{
    \centering
    \includegraphics[width=0.325\textwidth]{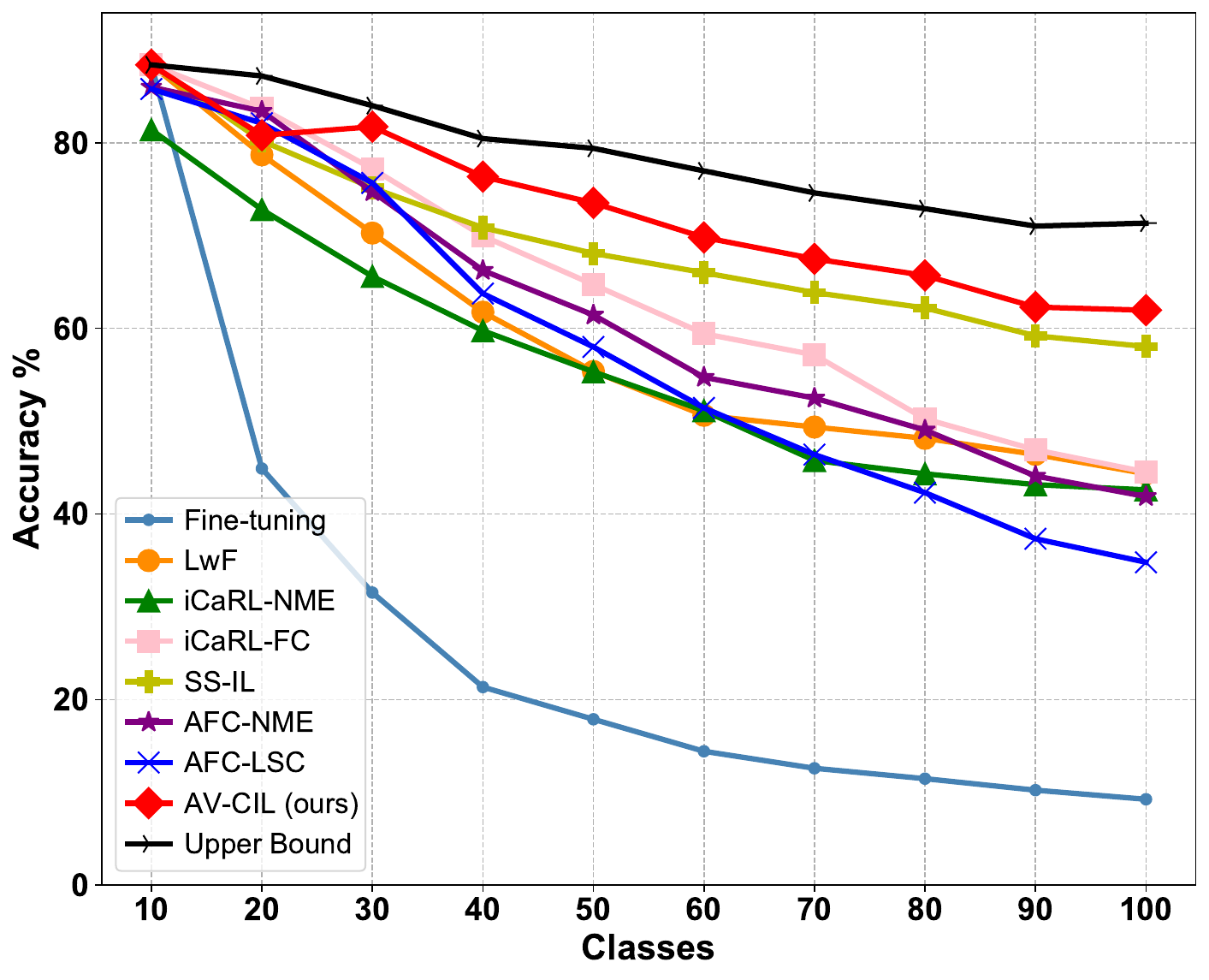}
    \label{fig:acc_VS_av}
    }
    \vspace{-3mm}
    \caption{Testing accuracy at each incremental step on (a) AVE-CI, (b) K-S-CI, and (c) VS100-CI datasets. The results show that as the incremental step increases,
    our AV-CIL generally outperforms other state-of-the-art incremental learning methods. }
\end{figure*}

\noindent
\textbf{Evaluation Metric.}
We evaluate the performance of all the methods in our experiments with Mean Accuracy and Average Forgetting, two common evaluation metrics in class-incremental learning. Mean Accuracy is the average of the testing accuracy of each task, which can be denoted as:
\begin{equation}
    \begin{split}
        MeanAcc. = \frac{1}{T}\sum_{t=1}^T a_t,
    \end{split}
    \label{eq:mean_acc}
\end{equation}
where $a_t$ denotes testing accuracy of all seen classes after completing the training on task $\mathcal{T}_t$ (incremental step $t$). Average Forgetting is used to measure the extent of catastrophic forgetting over previously learned tasks, which can be denoted as:
\begin{equation}
    \begin{split}
        Ave.Forget. &= \frac{1}{T-1}\sum_{t=2}^T F_t, \\
        \textit{s.t.}\,\,\, F_t = \frac{1}{t-1}\sum_{i=1}^{t-1} &\mathop{\text{max}}\limits_{\tau\in\{1,...,t-1\}}(a_{\tau,i}-a_{t,i}),
    \end{split}
\end{equation}
where $a_{\tau,i}$ is the testing accuracy of the $i$-th task after training on the $\tau$-th task (task $\mathcal{T}_\tau$ or incremental step $\tau$).

\noindent
\textbf{Implementation Details.} We conduct all our experiments with PyTorch~\cite{paszke2019pytorch}. For the data pre-processing, we follow the protocol of VideoMAE~\cite{tong2022videomae} and AudioMAE~\cite{huang2022masked}. We freeze the self-supervised pre-trained visual and audio encoder during the training of the baselines and our method, and fine-tune the rest parts of the model (audio-guided visual attention layer, audio-visual fusion layer, and the classifier). We use Adam~\cite{kingma2015adam} optimizer to train the model with learning rate and weight decay of 1e-3 and 1e-4, respectively. For all three datasets, we set the maximum training epochs number in each incremental step, the balance weight $\lambda_{VAD}$ in $\mathcal{L}_{VAD}$, and the temperature $\tau$ in D-AVSC to 200, 0.5, and 0.05, respectively. For the AVE-CI dataset, we set the memory size to 340. For the weights of losses $\lambda_I$ and $\lambda_C$, we set them to 0.5 and 1.0, respectively. For the K-S-CI dataset, we set the memory size, $\lambda_I$, and $\lambda_C$ to 500, 0.1, and 1.0, respectively. For the VS100-CI dataset, the memory size, each incremental step's training epoch, $\lambda_I$, and $\lambda_C$ are set to 1500, 200, 0.1, and 1.0, respectively.

\begin{figure*}
    \centering
    \includegraphics[width=0.98\textwidth]{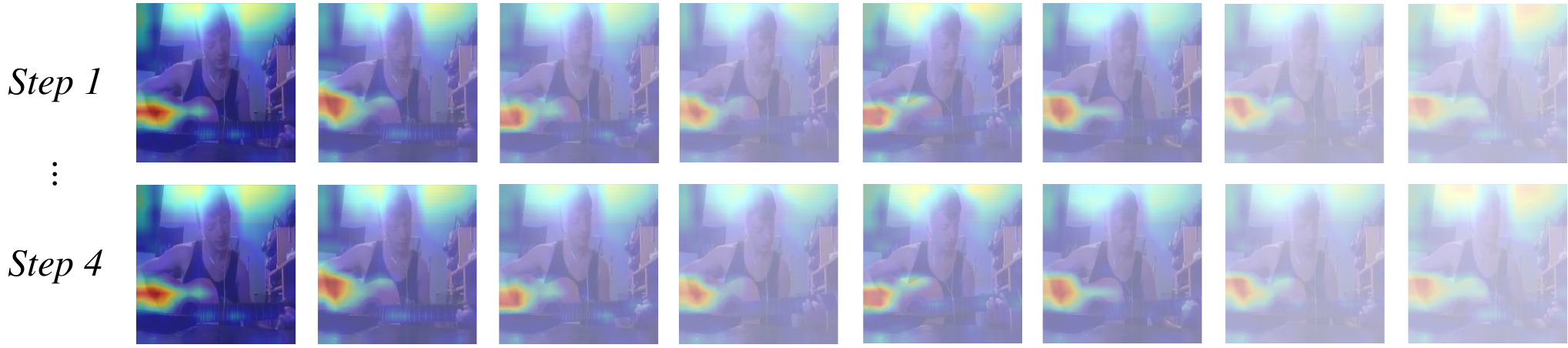}
    \caption{Visualization of the visual attention map as the incremental step increases with our proposed VAD. After the utilization of the VAD, the previously learned audio-guided visual attentive abilities, as well as the learned audio-visual correlations, are preserved by the model in new tasks.}
    \label{fig:attn_distill_after}
\end{figure*}

\subsection{Experimental Comparison}
We show our main experimental results in Table~\ref{tab:main_res}. We can see that our proposed AV-CIL outperforms recent state-of-the-art methods significantly. Specifically, on the AVE-CI dataset, our AV-CIL outperforms the state-of-the-art Mean Accuracy and Average Forgetting results by \textbf{5.58} and \textbf{4.08}, respectively. For K-S-CI, our method outperforms the state-of-the-art method by \textbf{3.35} and \textbf{4.05} for Mean Accuracy and Average Forgetting, respectively. For the VS100-CI dataset, our method has the improvement of \textbf{3.60} and \textbf{4.26} for Mean Accuracy and Average Forgetting over state-of-the-art results. These experimental results demonstrate the effectiveness of our proposed method in audio-visual class-incremental learning. We also show the testing accuracy at each incremental step of our AV-CIL, baselines, and the upper bound on the AVE-CI, K-S-CI, and VS100-CI datasets in Figure~\ref{fig:acc_AVE_av},~\ref{fig:acc_KS_av}, and~\ref{fig:acc_VS_av}. We can see that our method achieves the best performance at each incremental step on the AVE-CI dataset. For the K-S-CI and VS100-CI datasets, our proposed AV-CIL also has the best overall results and the best performance at each incremental step, except at step 2. In summary, our method has a significant superiority in audio-visual class-incremental learning compared to state-of-the-art class-incremental learning methods, which demonstrates the effectiveness of the combination of our proposed D-AVSC and VAD.

\begin{table*}[htbp]
  \centering
  \caption{Ablation study on input modalities. We can see that joint audio-visual modeling is better than individual modeling in class-incremental learning.}
    \begin{tabular}{lccccccccc}
    \toprule
    \multirow{3}[4]{*}{Methods} & \multicolumn{9}{c}{Mean Accuracy} \\
    & \multicolumn{3}{c}{AVE-CI} & \multicolumn{3}{c}{K-S-CI} & \multicolumn{3}{c}{VS100-CI} \\ \cmidrule(r){2-4} \cmidrule(r){5-7} \cmidrule(r){8-10}  & Audio & Visual & Aud.-Vis. & Audio & Visual & Aud.-Vis. & Audio & Visual & Aud.-Vis. \\
    \midrule
    Fine-tuning & 38.41 & 35.68 & 42.40  & 32.60  & 38.49 & 41.18 & 24.40 & 23.06 & 26.21 \\
    LwF~\cite{li2017learning}  & 51.88 & 50.09 & 58.07 & 47.68 & 61.91 & 65.54 & 49.49 & 49.10 & 59.34 \\
    iCaRL-NME~\cite{rebuffi2017icarl} & 44.62 & 52.72 & 56.15 & 39.75 & 61.47 & 64.51 & 41.61 & 46.61 & 56.62 \\
    iCaRL-FC~\cite{rebuffi2017icarl} & 56.17 & 55.06 & 65.88 & 43.14 & 58.74 & 65.05 & 54.46 & 48.39 & 64.22 \\
    SS-IL~\cite{ahn2021ssil} & 53.81 & 52.10  & 61.94 & 47.65 & 64.55 & 69.71 & 57.88 & 53.86 & 69.20 \\
    AFC-NME~\cite{kang2022class} & 55.28 & 58.67 & 68.46 & 46.71 & 66.63 & 69.13 & 49.91 & 54.50 & 61.10 \\
    AFC-LSC~\cite{kang2022class} & 56.36 & 57.92 & 65.21 & 39.89 & 64.88 & 67.02 & 44.82 & 51.69 & 58.15 \\
    \midrule
    Oracle (Upper Bound) & 63.66 & 61.55 & 76.85 & 53.39 & 71.94 & 80.43 & 66.22 & 61.75 & 78.63 \\
    \bottomrule
    \end{tabular}%
  \label{tab:sing_modality}%
\end{table*}%

\begin{table}[htbp]
  \centering
  \caption{Ablation study on our AV-CIL. Our full model can better handle catastrophic forgetting and achieve the best incremental learning performance on AVE-CI dataset.}
    \scalebox{0.9}{\begin{tabular}{ccccc}
    \toprule
    \multirow{10}[6]{*}{AV-CIL} & \multicolumn{2}{c}{D-AVSC} & \multirow{2}[4]{*}{VAD} & \multirow{2}[4]{*}{Mean Acc.} \\
\cmidrule{2-3}          & I-AVSS & C-AVSS &       &  \\
\cmidrule{2-5}          & \XSolidBrush     & \XSolidBrush     & \XSolidBrush     & 62.15 \\
          & \CheckmarkBold     & \XSolidBrush     & \XSolidBrush     & 68.38 \\
          & \XSolidBrush     & \CheckmarkBold     & \XSolidBrush     & 63.81 \\
          & \XSolidBrush     & \XSolidBrush     & \CheckmarkBold     & 63.49 \\
          & \CheckmarkBold     & \CheckmarkBold     & \XSolidBrush     & 69.01 \\
          & \CheckmarkBold     & \XSolidBrush     & \CheckmarkBold     & 72.35 \\
          & \XSolidBrush     & \CheckmarkBold     & \CheckmarkBold     & 63.95 \\
          & \CheckmarkBold     & \CheckmarkBold     & \CheckmarkBold     & 74.04 \\
    \bottomrule
    \end{tabular}%
    }
  \label{tab:ablation}%
\end{table}%

\subsection{Ablation Study}
Our proposed method AV-CIL mainly contributes two parts: Dual-Audio-Visual Similarity Constraint (D-AVSC) that consists of the Instance-aware Audio-Visual Semantic Similarity (I-AVSS) and the Class-aware Audio-Visual Semantic Similarity (C-AVSS), and Visual Attention Distillation (VAD). In this subsection, we conduct the ablation study of our method by removing single or multiple parts from D-AVSC and VAD when running experiments. The results are presented in Table~\ref{tab:ablation}, from which we can see that each of our proposed parts has a positive impact on the final results. This demonstrates the effectiveness of our proposed D-AVSC (including both I-AVSS and D-AVSS) and VAD.

\subsection{Comparison with Single Modality Results}
To evaluate the effectiveness of class-incremental learning with joint audio-visual modalities compared to using only the audio or visual modality, in this subsection, we conduct experiments of class-incremental learning with only single audio or visual modality and compare them with our experimental results with joint audio-visual modalities. We show the results in Table~\ref{tab:sing_modality}, from which we can see that, for all the baselines, compared to training with single audio or visual modality, training with audio-visual modalities achieves higher Mean Accuracy in class-incremental learning, which demonstrates that joint audio-visual modeling is better for class-incremental learning compared to only using data from single audio or visual modality for class-incremental learning. Please see Appendix for the comparison at each step. 

\begin{figure}[t]
    \centering
    \subfloat[]{
    \centering
    \includegraphics[width=0.23\textwidth]{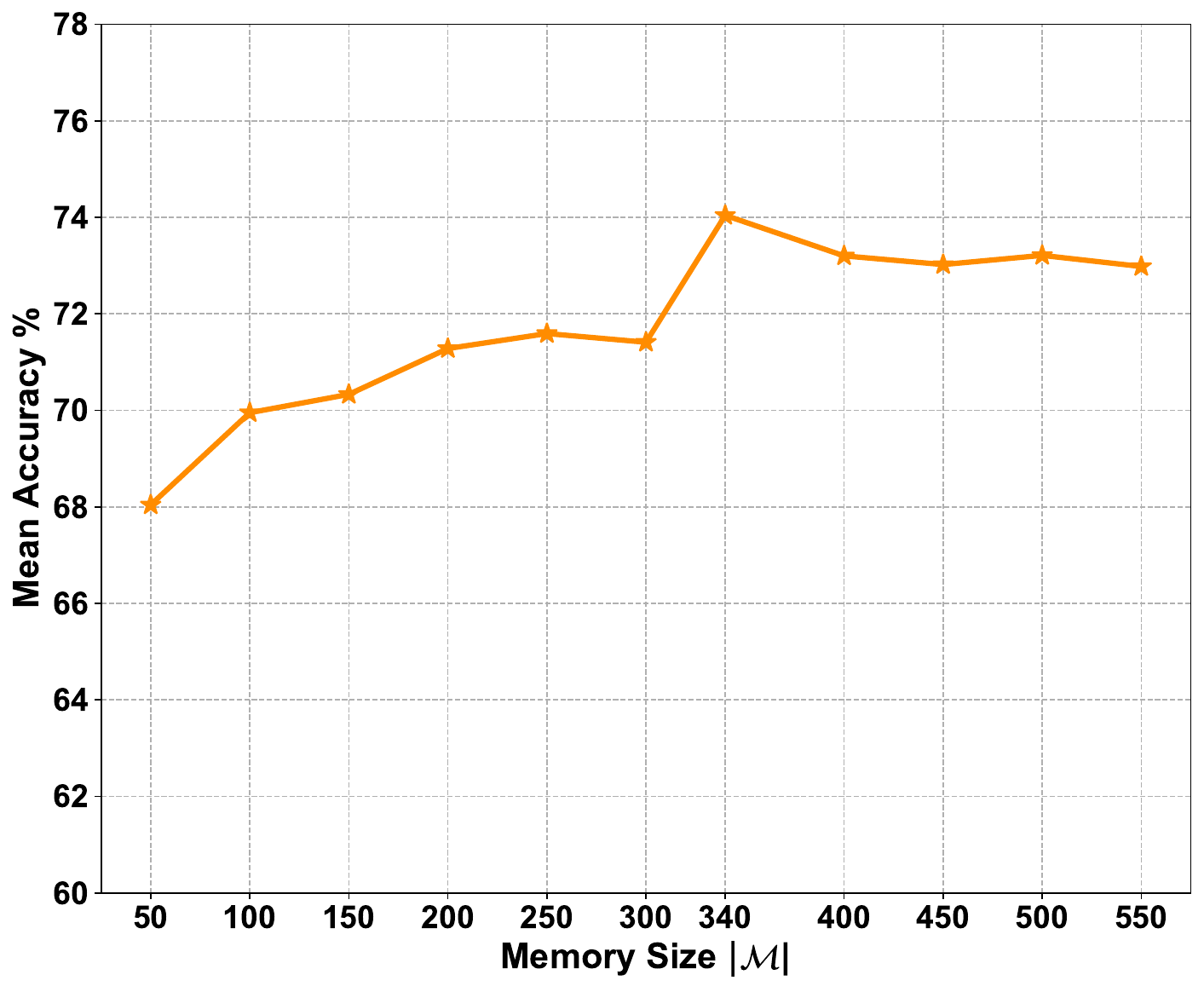}
    \label{fig:memory_AVE}
    }
    \subfloat[]{
    \centering
    \includegraphics[width=0.23\textwidth]{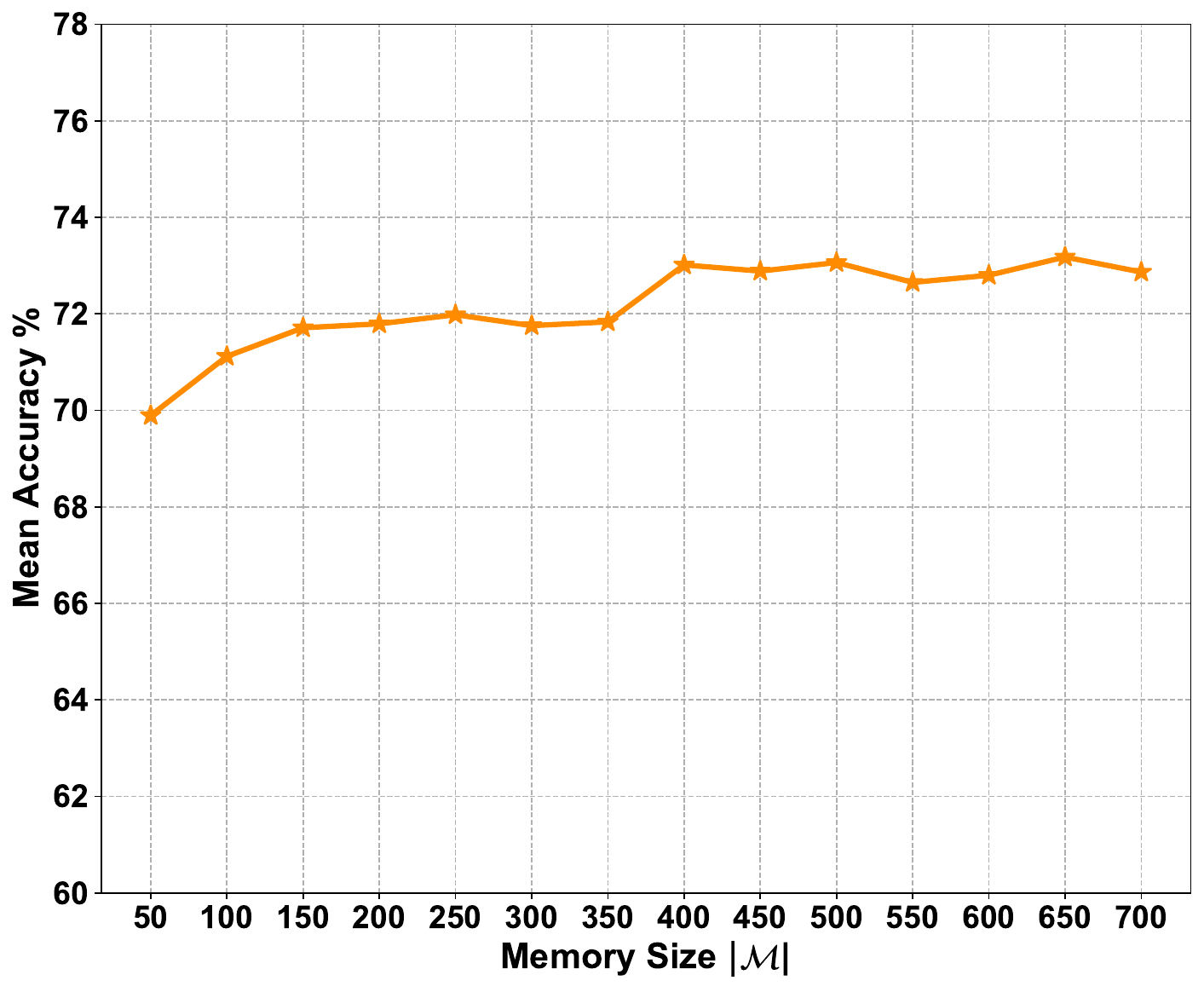}
    \label{fig:memory_KS}
    }
    \caption{Experimental results of our proposed AV-CIL with different memory size ($|\mathcal{M}|$) on (a) AVE-CI, and (b) K-S-CI dataset.}
\end{figure}

\subsection{Effect of Memory Size}
In this section, we investigate the effect of memory size on our proposed class-incremental learning approach. We conduct experiments with different memory sizes and report the results in Figure~\ref{fig:memory_AVE} and~\ref{fig:memory_KS} for AVE-CI and K-S-CI datasets, respectively. 
Our experimental results show that the performance of our model generally improves followed by stable fluctuations as the memory size increases on both datasets. Specifically, on the AVE-CI dataset, the model's performance consistently improves as the memory size grows from 50 to 340. However, a fluctuation in the performance is following as the memory size increases from 340 to 550. Similarly, on the K-S-CI dataset, the model's performance tends to increase as the memory size grows from 50 to 400, followed by a stable fluctuation as the memory size increases from 400 to 700. These trends suggest that the model's performance steadily improves in the early stages of memory growth, and then reaches a plateau where further increases in memory size do not lead to significant performance improvements.


\subsection{Visualization of Visual Attention Map}
In Figure~\ref{fig:attn_distill_after}, we show the visualization of the visual attention map after applying our proposed Visual Attention Distillation (VAD). The illustrated sample is the same as that in Figure~\ref{fig:attn_distill_motivation}. We can see that the attentive abilities learned in previous tasks can be preserved. This demonstrates that our proposed VAD can prevent the model from forgetting the learned audio-visual correlations in previous classes. For the full version of Figure~\ref{fig:attn_distill_after}, please see Appendix for details.

\section{Conclusion}
In this paper, we explore the effectiveness of training with joint audio-visual modalities in the alleviation of the catastrophic forgetting problem in class-incremental learning and propose the audio-visual class-incremental learning, a novel class-incremental learning problem with audio-visual data. To tackle our proposed problem, we propose a method named AV-CIL, which contains the Dual-Audio-Visual Similarity Constraint to preserve the cross-modal similarity from the perspective of both instance and class. Moreover, to preserve the learned semantic correlations between audio and visual modalities as the incremental step increases, we propose Visual Attention Distillation to distill the learned audio-guided visual attentive ability into new incremental steps. Experimental results on three constructed audio-visual class-incremental datasets AVE-CI, K-S-CI, and VS100-CI show that our proposed approach outperforms state-of-the-art methods significantly. This paper provides a new direction in class-incremental learning.

\vspace{2mm}
\noindent
 \textbf{Acknowledgments.} 
We would like to thank the anonymous reviewers for their constructive comments. This work was supported in part by gifts from Cisco Systems and Adobe. The article solely reflects the opinions and conclusions of its authors but not the funding agents.

{\small
\bibliographystyle{ieee_fullname}
\bibliography{egbib}
}

\appendix

\section{Appendix}
In this appendix, we present supplementary experimental results for various class-incremental learning approaches utilizing distinct input modalities at each step, as detailed in Section \ref{sec:perstepunimodal}. Following that, we offer comprehensive visualizations of visual attention maps in Section \ref{sec:att}. 
Moreover, in Section \ref{sec:param}, we supply further parameter studies, and in Section~\ref{sec:setting}, we include additional results under diverse class-incremental settings. Finally, we show the experimental results compared to the existing visual attention distillation approach in Section~\ref{sec:visual_attn_distil}.

\begin{table}[htbp]
  \centering
  \caption{Parameter studies on AVE-CI dataset.}
  \resizebox{0.48\textwidth}{!}{
    \begin{tabular}{cc|cccc|c}
    \toprule
    \multirow{2}[2]{*}{$\lambda_I$} & \multirow{2}[2]{*}{$\lambda_C$} & \multicolumn{4}{c}{Accuracy}      & \multirow{2}[2]{*}{Mean Acc.} \\
          &       & step 1 & step 2 & step 3 & \multicolumn{1}{c}{step 4} &  \\
    \midrule
    0.1   & 1.0     & 79.81 & 71.43 & 69.52 & 66.50  & 71.82 \\
    0.3   & 1.0     & 79.81 & 75.71 & 71.75 & 67.26 & 73.63 \\
    0.5   & 1.0     & 79.81 & 77.14 & 71.43 & 67.77 & 74.04 \\
    0.8   & 1.0     & 79.81 & 76.19 & 70.16 & 65.23 & 72.85 \\
    1.0   & 1.0     & 79.81 & 75.24 & 67.94 & 66.50  & 72.37 \\
    0.5   & 0.8   & 79.81 & 75.24 & 71.43 & 68.78 & 73.82 \\
    0.5   & 0.5   & 79.81 & 74.29 & 67.94 & 66.24 & 72.07 \\
    0.5   & 0.3   & 79.81 & 75.71 & 70.48 & 68.27 & 73.56 \\
    0.5   & 0.1   & 79.81 & 75.71 & 68.25 & 66.50 & 72.57 \\
    \bottomrule
    \end{tabular}%
    }
  \label{tab:parameter_study_AVE}%
\end{table}%

\begin{table}[htbp]
  \centering
  \caption{Parameter studies on K-S-CI dataset.}
  \resizebox{0.48\textwidth}{!}{
    \begin{tabular}{cc|ccccc|c}
    \toprule
    \multirow{2}[2]{*}{$\lambda_I$} & \multirow{2}[2]{*}{$\lambda_C$} & \multicolumn{5}{c}{Accuracy}      & \multirow{2}[2]{*}{Mean Acc.} \\
          &       & step 1 & step 2 & step 3 & step 4 & \multicolumn{1}{c}{step 5} &  \\
    \midrule
    0.1   & 1.0     & 93.01 & 73.64 & 70.19 & 66.47 & 62.00 & 73.06 \\
    0.3   & 1.0     & 93.01 & 70.13 & 67.76 & 64.40 & 60.32 & 71.12 \\
    0.5   & 1.0     & 93.01 & 70.26 & 66.64 & 63.56 & 58.53 & 70.40 \\
    0.8   & 1.0     & 93.01 & 66.10 & 63.78 & 61.55 & 57.05 & 68.30 \\
    1.0   & 1.0     & 93.01 & 68.05 & 65.08 & 62.01 & 56.03 & 68.84 \\
    0.1   & 0.8     & 93.01 & 72.99 & 69.50 & 66.73 & 61.64 & 72.77 \\
    0.1   & 0.5     & 93.01 & 72.47 & 68.63 & 66.21 & 60.88 & 72.24 \\
    0.1   & 0.3     & 93.01 & 73.38 & 69.41 & 65.44 & 61.44 & 72.54 \\
    0.1   & 0.1     & 93.01 & 71.95 & 68.80 & 66.34 & 60.27 & 72.07 \\
    \bottomrule
    \end{tabular}%
    }
  \label{tab:parameter_study_KS}%
  \vspace{-10pt}
\end{table}%

\begin{table*}[t]
  \centering
  \caption{Parameter studies on VS100-CI dataset.}
    \begin{tabular}{cc|cccccccccc|c}
    \toprule
    \multirow{2}[2]{*}{$\lambda_I$} & \multirow{2}[2]{*}{$\lambda_C$} & \multicolumn{10}{c}{Accuracy}      & \multirow{2}[2]{*}{Mean Acc.} \\
          &       & step 1 & step 2 & step 3 & step 4 & step 5 & step 6 & step 7 & step 8 & step 9 & \multicolumn{1}{c}{step 10} &  \\
    \midrule
    0.1   & 1.0 & 88.40 & 80.80 & 81.73 & 76.35 & 73.52 & 69.80 & 67.49 & 65.70 & 62.29 & 61.96 & 72.80 \\
    0.3   & 1.0  & 88.40 & 78.40 & 79.13 & 75.25 & 71.84 & 67.57 & 65.97 & 63.83 & 60.31 & 60.26 & 71.10 \\
    0.5   & 1.0 & 88.40 & 77.30 & 77.27 & 76.00 & 70.96 & 67.07 & 65.49 & 62.03 & 58.09 & 57.40 & 70.00 \\
    0.8  & 1.0 & 88.40 & 75.90 & 76.33 & 75.30 & 69.72 & 66.67 & 64.54 & 61.60 & 57.56 & 56.40 & 69.24 \\
    1.0  & 1.0 & 88.40 & 75.30 & 76.00 & 74.65 & 70.28 & 66.33 & 64.20 & 60.85 & 57.22 & 55.46 & 68.87 \\
    0.1 & 0.8 & 88.40 & 81.30 & 81.27 & 76.55 & 73.32 & 69.23 & 68.00 & 65.40 & 62.20 & 61.62 & 72.73 \\
    0.1  & 0.5  & 88.40 & 81.20 & 81.67 & 76.60 & 72.52 & 69.47 & 67.46 & 65.08 & 61.96 & 61.88 & 72.62 \\
    0.1   & 0.3 & 88.40 & 80.40 & 81.07 & 76.20 & 72.88 & 69.27 & 66.69 & 65.05 & 62.18 & 61.54 & 72.37 \\
    0.1  & 0.1 & 88.40 & 80.40 & 81.07 & 75.80 & 72.32 & 68.83 & 67.26 & 65.15 & 61.98 & 61.30 & 72.25  \\
    \bottomrule
    \end{tabular}%
  \label{tab:parameter_study_VS}%
\end{table*}%

\begin{table*}[t]
  \centering
  \caption{Experimental results of different methods on AVE-CI and K-S-CI datasets with different class-incremental settings.}
    \begin{tabular}{lcccc}
    \toprule
    \multirow{4}[2]{*}{Methods} & \multicolumn{4}{c}{Mean Accuracy} \\
          & \multicolumn{2}{c}{\multirow{2}[0]{*}{AVE-CI}} & \multicolumn{2}{c}{\multirow{2}[0]{*}{K-S-CI}} \\
          & \multicolumn{2}{c}{} & \multicolumn{2}{c}{} \\
          & 7 classes $\times$ 4 steps & 4 classes $\times$ 7 steps & 6 classes $\times$ 5 steps & 5 classes $\times$ 6 steps \\
    \midrule
    Fine-tuning & 42.40  & 37.54 & 41.18 & 37.45 \\
    LwF~\cite{li2017learning}   & 58.07 & 50.25 & 65.54 & 63.10 \\
    iCaRL-NME~\cite{rebuffi2017icarl} & 56.15 & 60.02 & 64.51 & 63.29 \\
    iCaRL-FC~\cite{rebuffi2017icarl} & 65.88 & 65.50  & 65.54 & 65.13 \\
    SS-IL~\cite{ahn2021ssil} & 61.94 & 65.29 & 69.71 & 68.38 \\
    AFC-NME~\cite{kang2022class} & 68.46 & 68.61 & 69.13 & 67.74 \\
    AFC-LSC~\cite{kang2022class} & 65.21 & 61.93 & 67.02 & 65.86 \\
    AV-CIL (Ours) & \textbf{74.04} & \textbf{71.76} & \textbf{73.06} & \textbf{73.24} \\
    \midrule
    Oracle (Upper Bound) & 76.85 & 78.23 & 80.43 & 80.50 \\
    \bottomrule
    \end{tabular}%
  \label{tab:diff_step_setting}%
\end{table*}%

\subsection{Effect of Input Modalities on Class-incremental Approaches}
\label{sec:perstepunimodal}
\begin{figure*}[t]
    \centering
    \subfloat[]{
    \centering
    \includegraphics[width=0.24\textwidth]{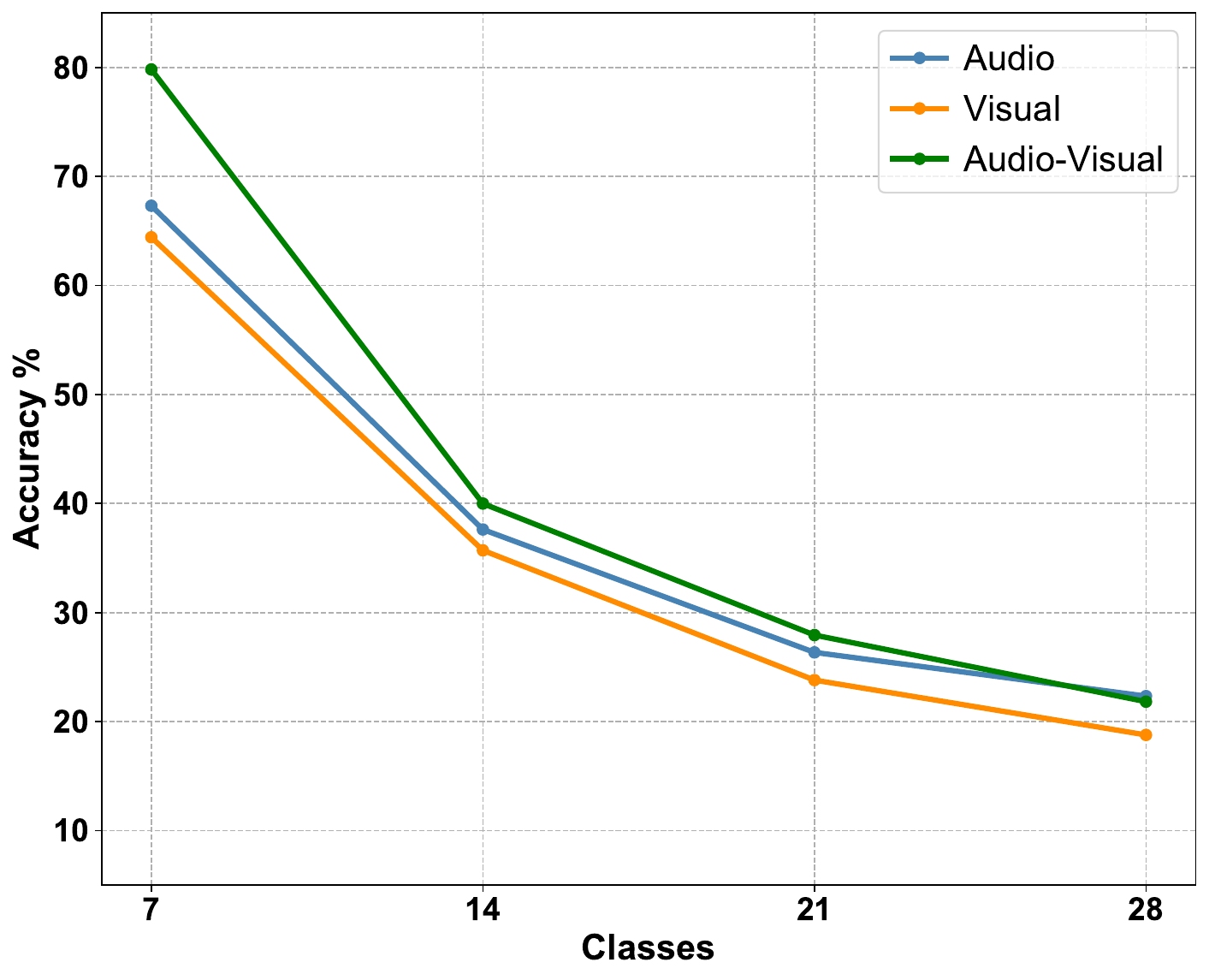}
    \label{fig:acc_AVE_fine_tuning}
    }
    \hspace{-6pt}
    \subfloat[]{
    \centering
    \includegraphics[width=0.24\textwidth]{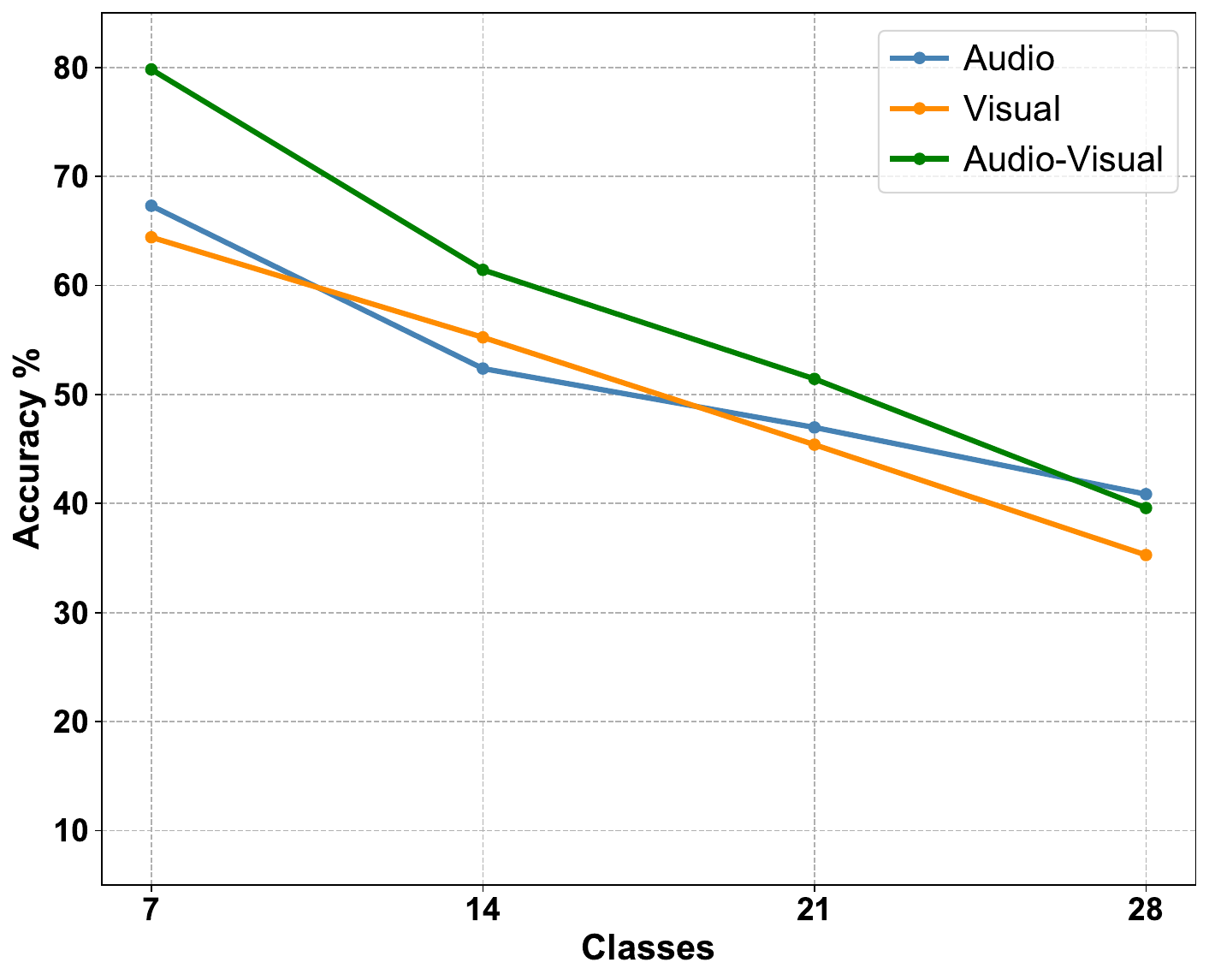}
    \label{fig:acc_AVE_lwf}
    }
    \hspace{-6pt}
    \subfloat[]{
    \centering
    \includegraphics[width=0.24\textwidth]{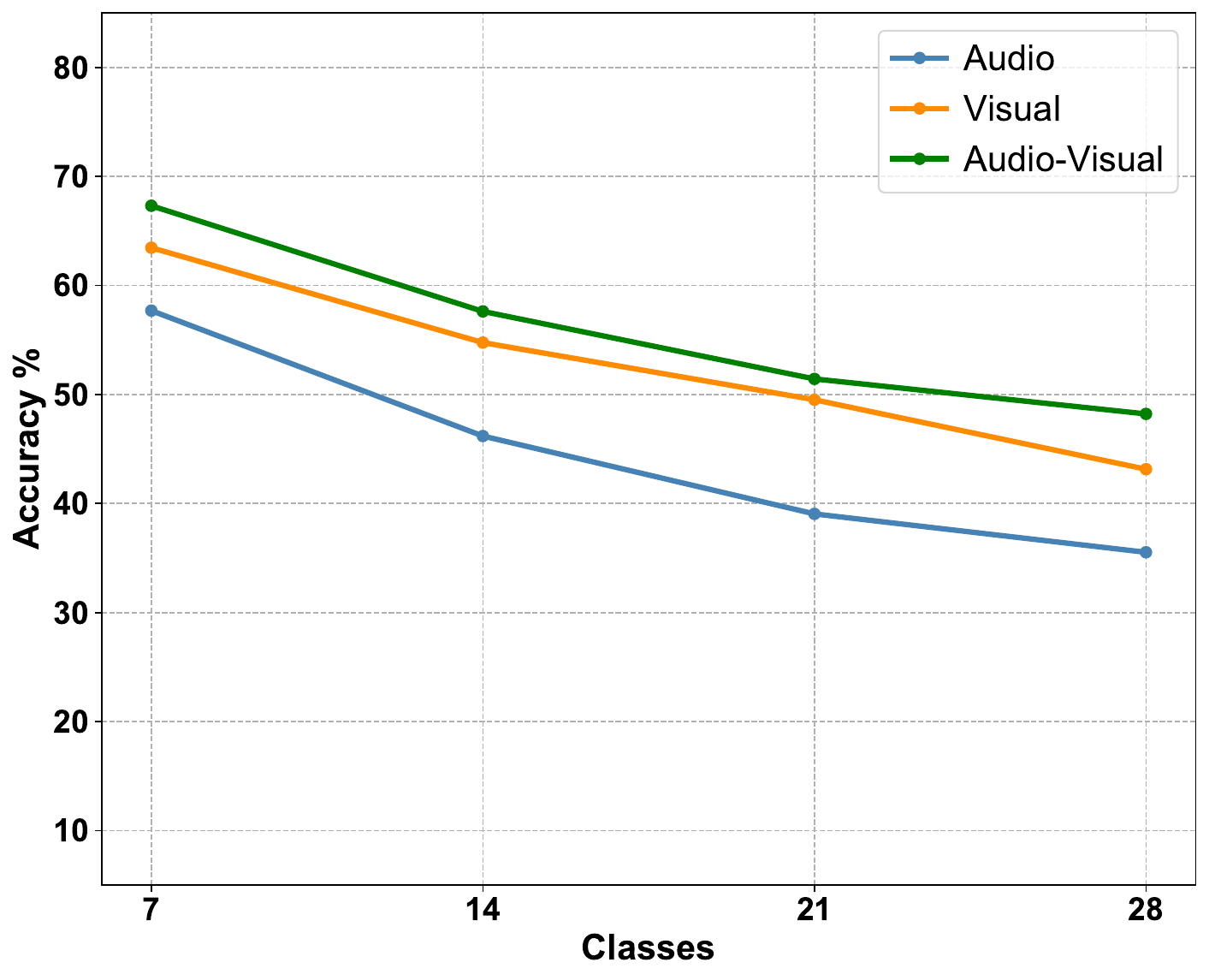}
    \label{fig:acc_AVE_iCaRL_NME}
    }
    \hspace{-6pt}
    \subfloat[]{
    \centering
    \includegraphics[width=0.24\textwidth]{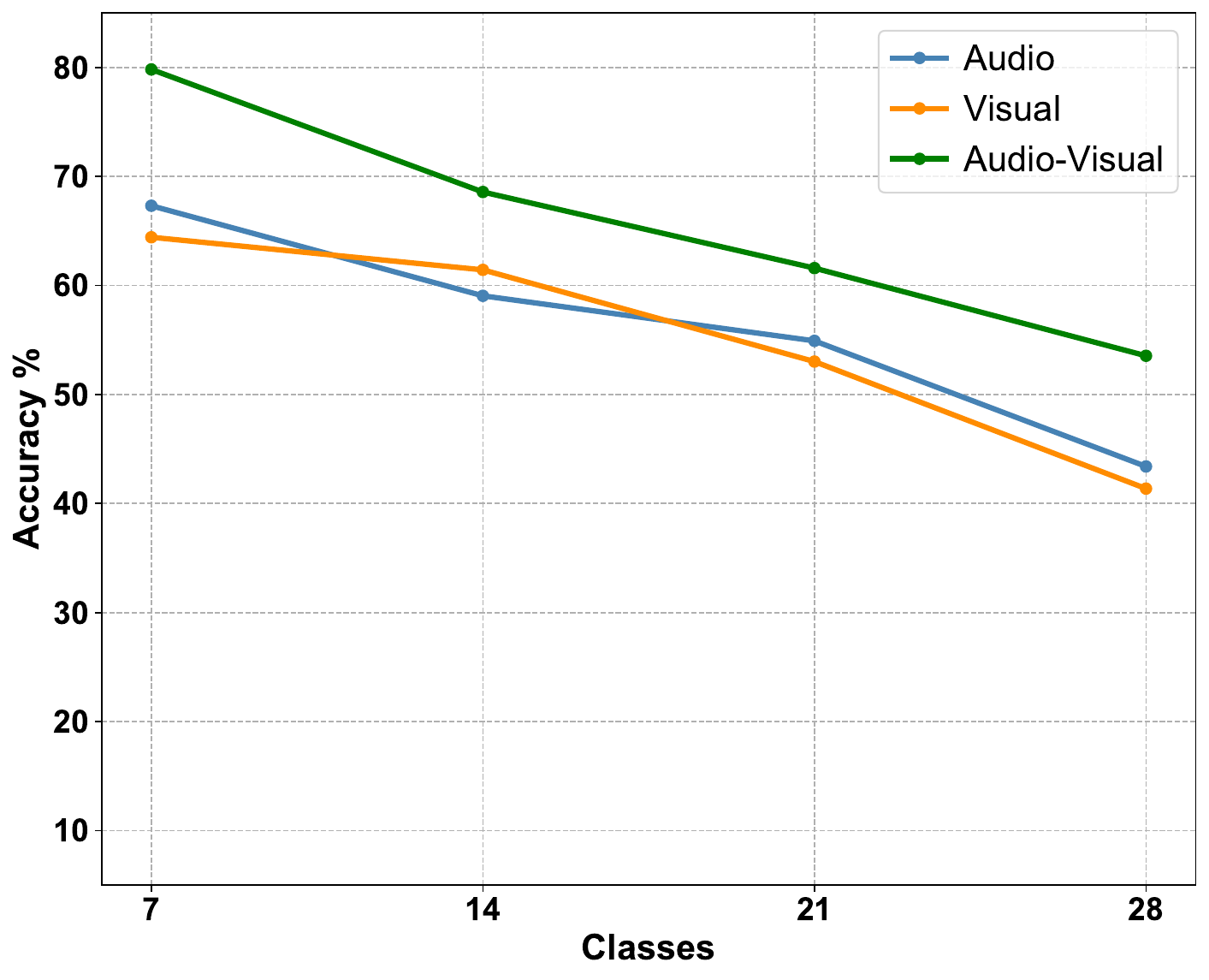}
    \label{fig:acc_AVE_iCaRL_FC}
    }
    \\
    \subfloat[]{
    \centering
    \includegraphics[width=0.24\textwidth]{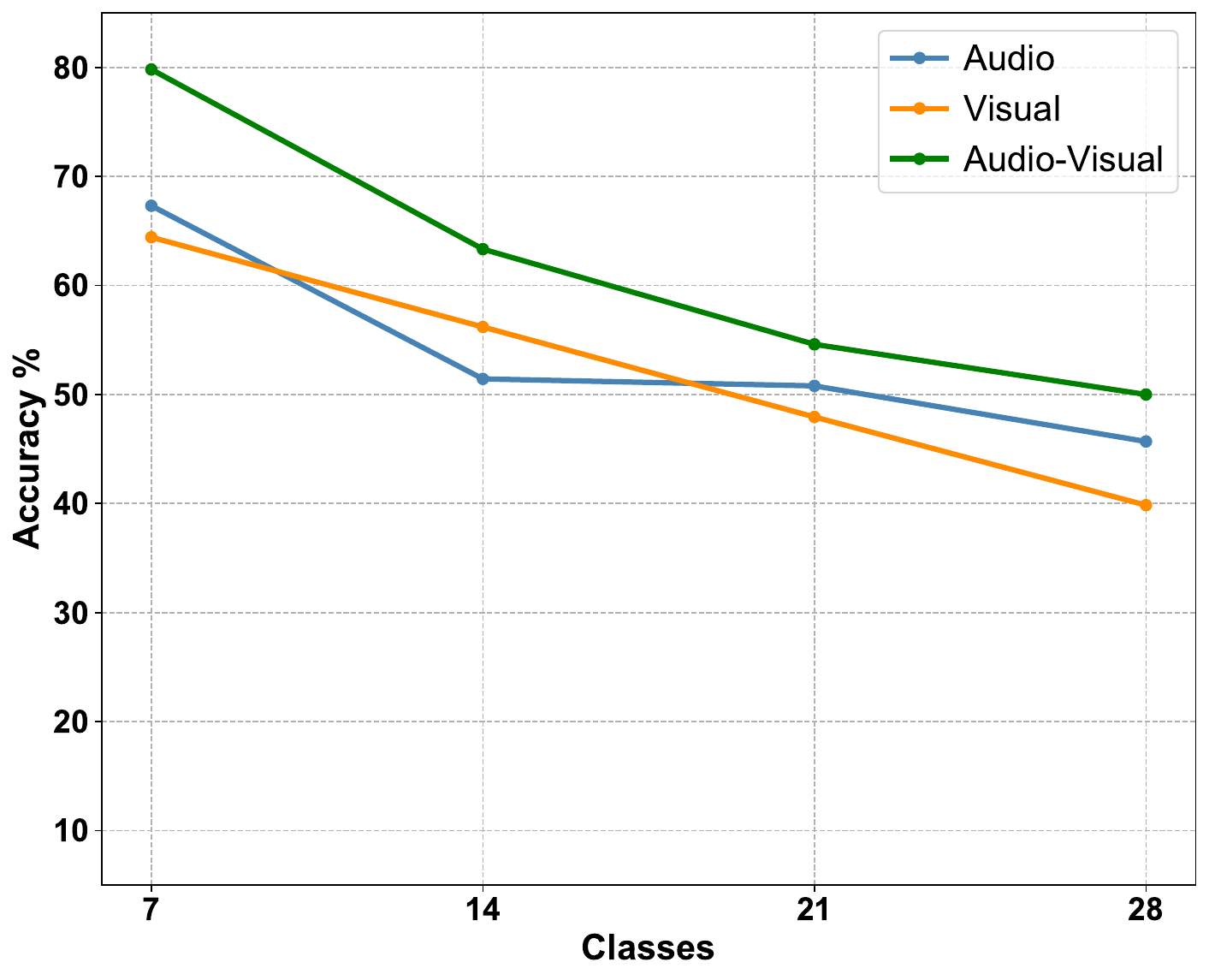}
    \label{fig:acc_AVE_SSIL}
    }
    \hspace{-6pt}
    \subfloat[]{
    \centering
    \includegraphics[width=0.24\textwidth]{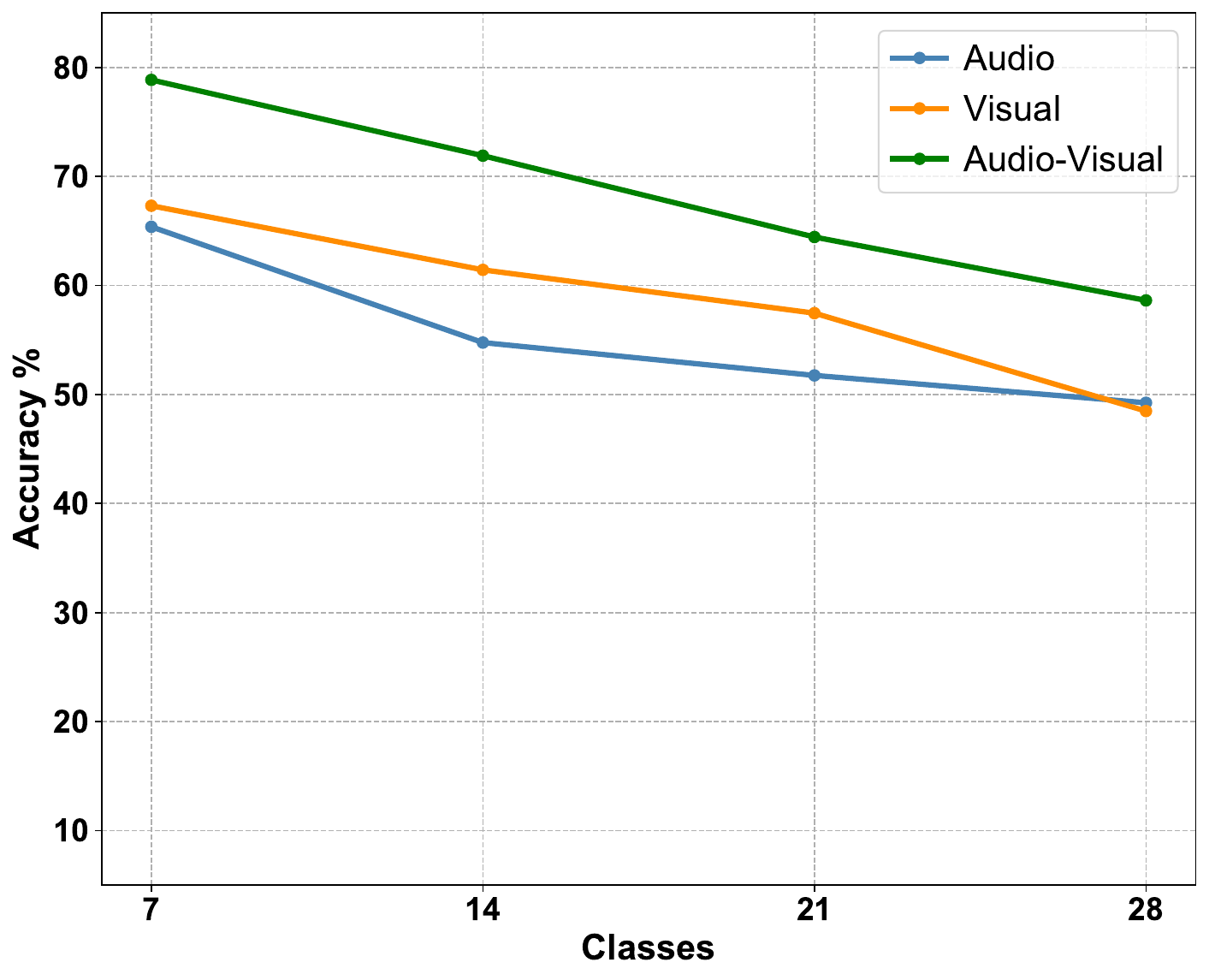}
    \label{fig:acc_AVE_AFC_NME}
    }
    \hspace{-6pt}
    \subfloat[]{
    \centering
    \includegraphics[width=0.24\textwidth]{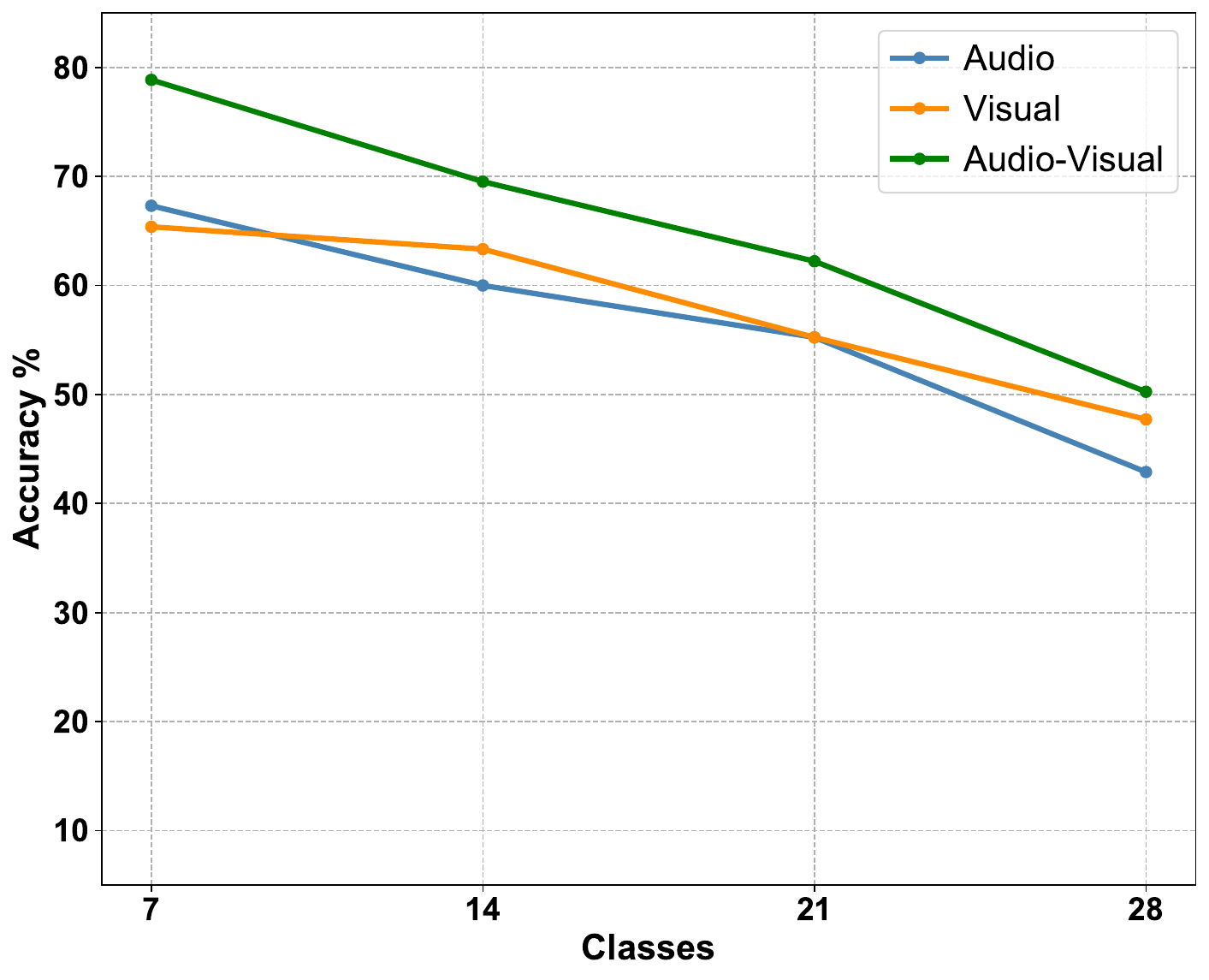}
    \label{fig:acc_AVE_AFC_LSC}
    }
    \hspace{-6pt}
    \subfloat[]{
    \centering
    \includegraphics[width=0.24\textwidth]{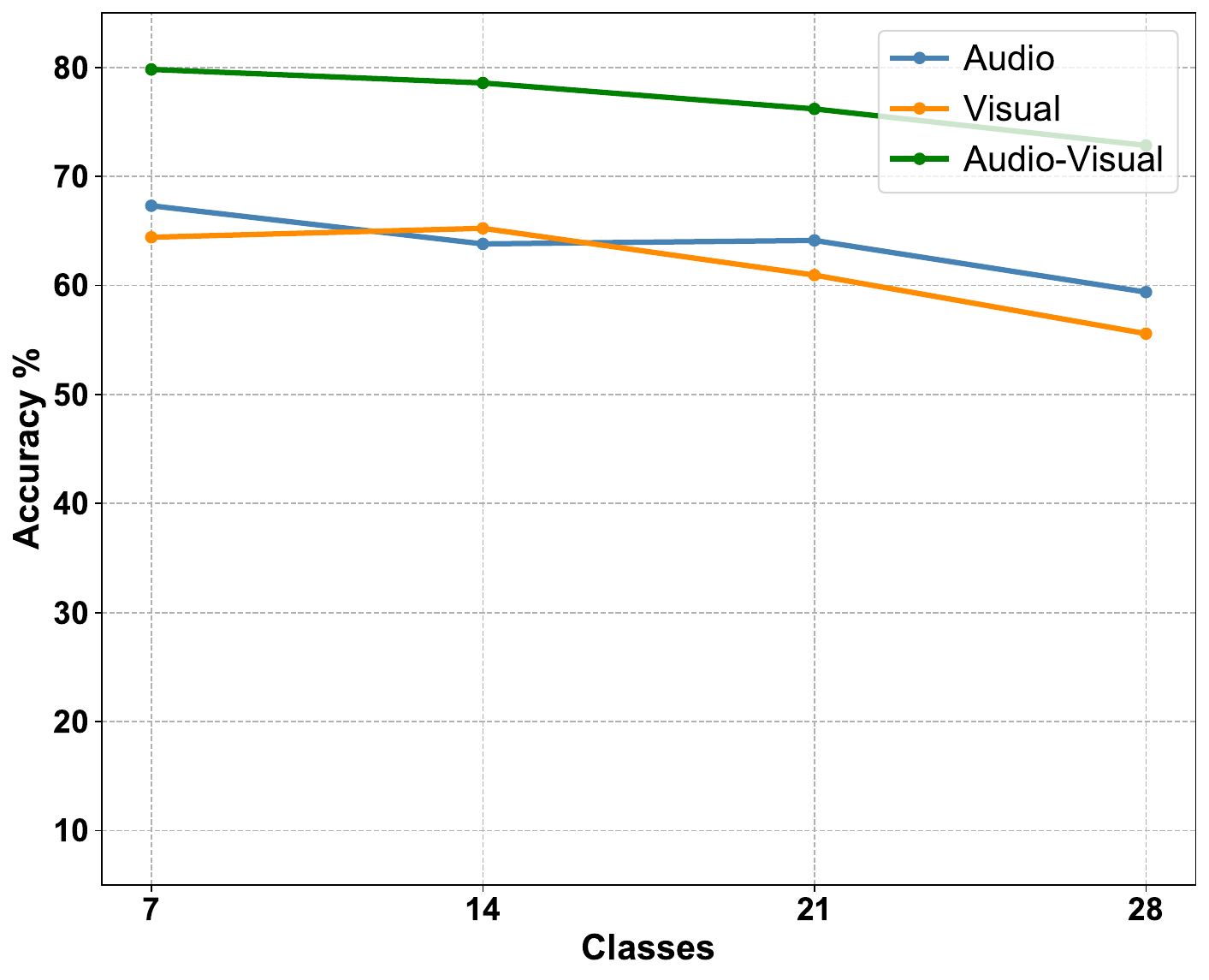}
    \label{fig:acc_AVE_upper}
    }
    \caption{Testing accuracy of training with audio, visual and audio-visual modalities of (a) Fine-tuning, (b) LwF, (c) iCaRL-NME, (d) iCaRL-FC, (e) SS-IL, (f) AFC-NME, (g) AFC-LSC, and (h) upper bound on AVE-CI dataset. 
    }
    \label{fig:acc_each_step_baselines_AVE}
\end{figure*}

\begin{figure*}[t]
    \centering
    \subfloat[]{
    \centering
    \includegraphics[width=0.24\textwidth]{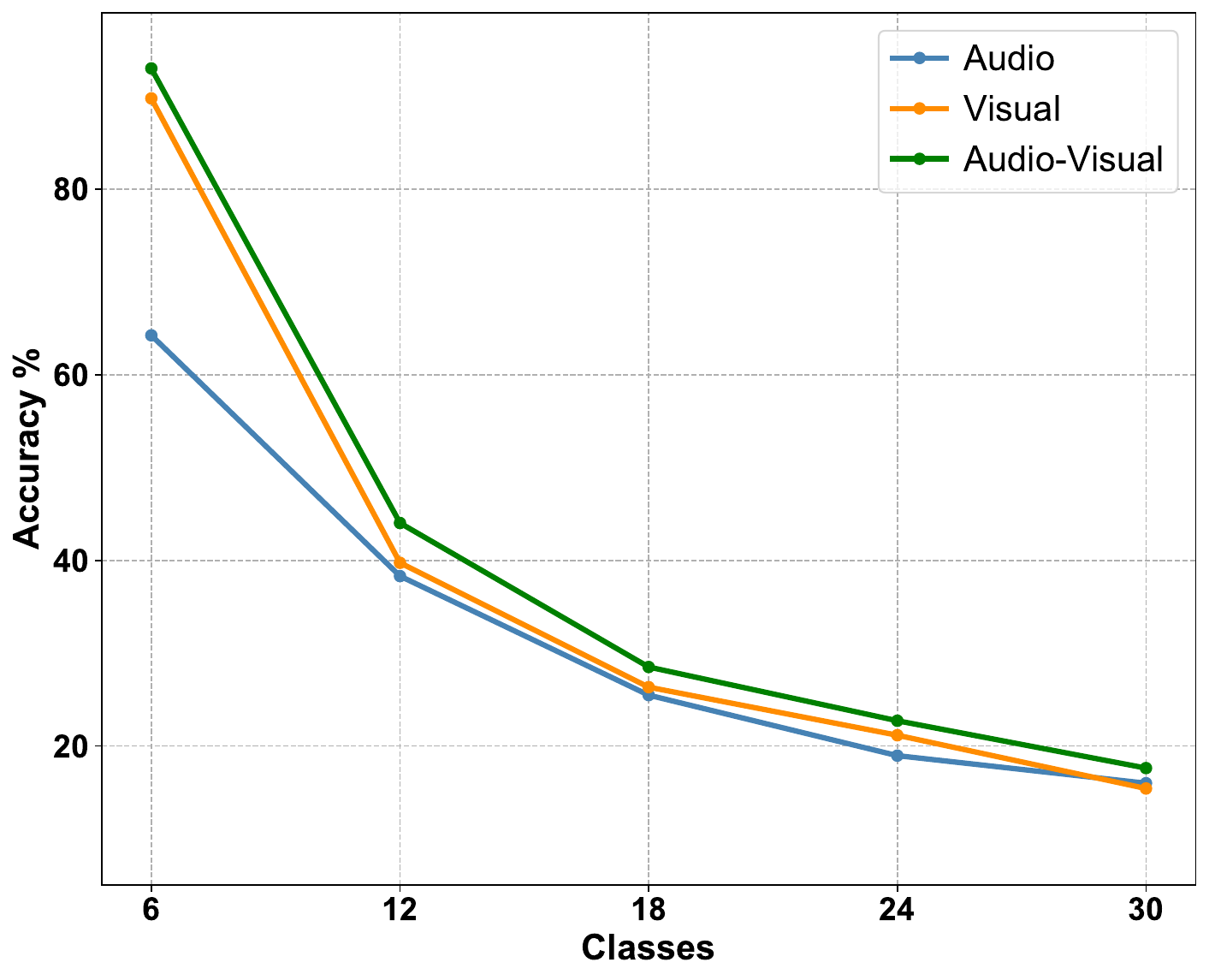}
    \label{fig:acc_KS_fine_tuning}
    }
    \hspace{-6pt}
    \subfloat[]{
    \centering
    \includegraphics[width=0.24\textwidth]{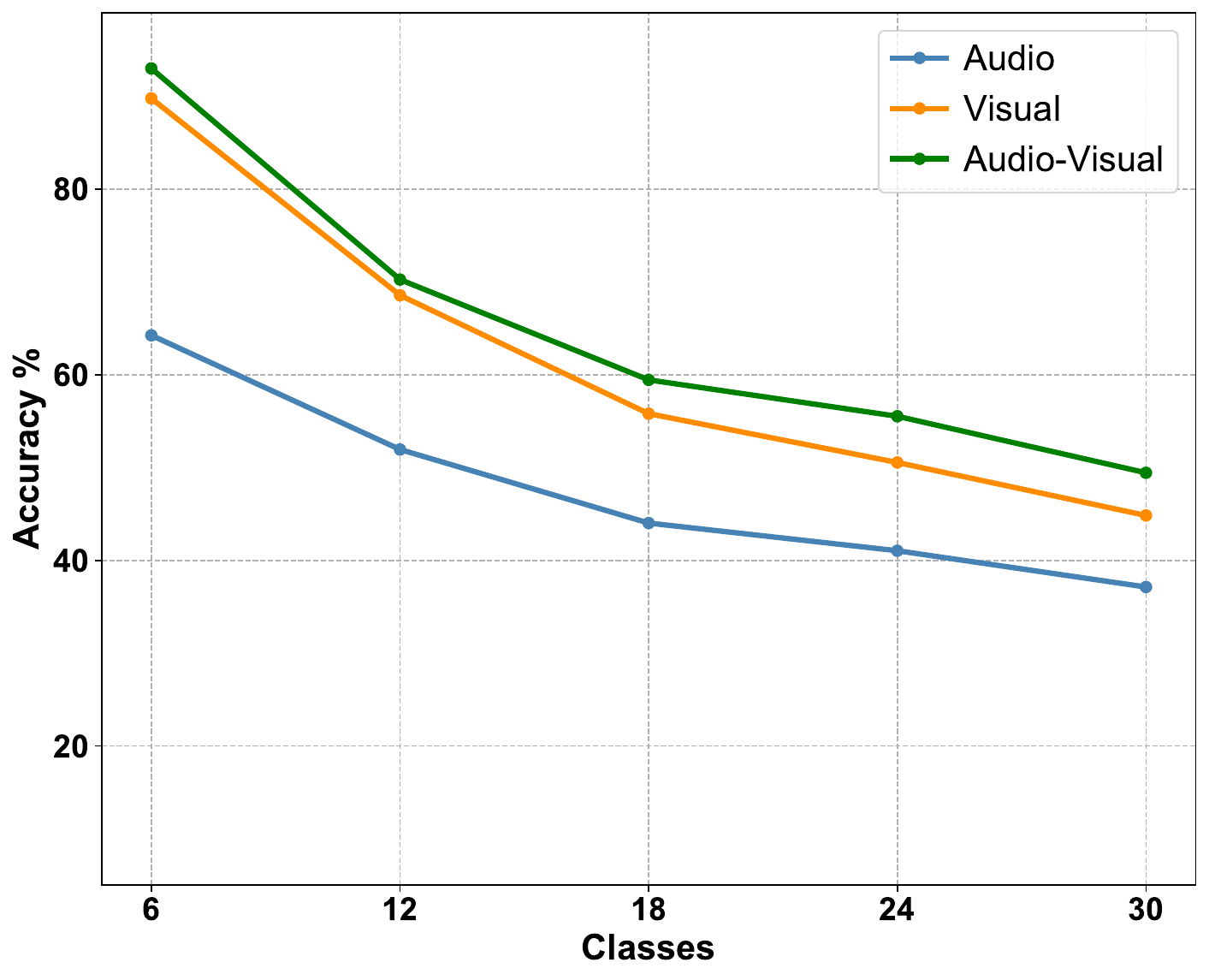}
    \label{fig:acc_KS_lwf}
    }
    \hspace{-6pt}
    \subfloat[]{
    \centering
    \includegraphics[width=0.24\textwidth]{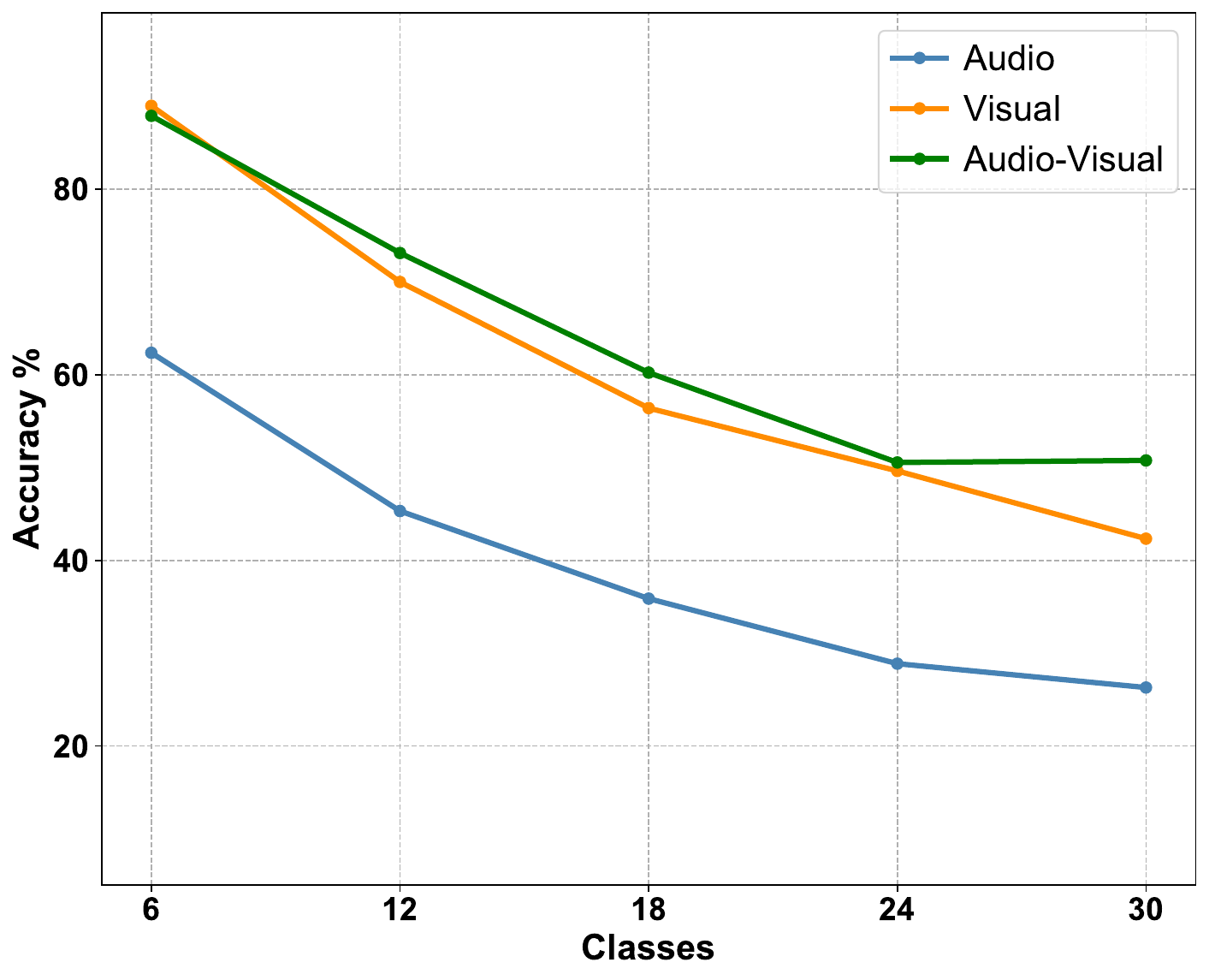}
    \label{fig:acc_KS_iCaRL_NME}
    }
    \hspace{-6pt}
    \subfloat[]{
    \centering
    \includegraphics[width=0.24\textwidth]{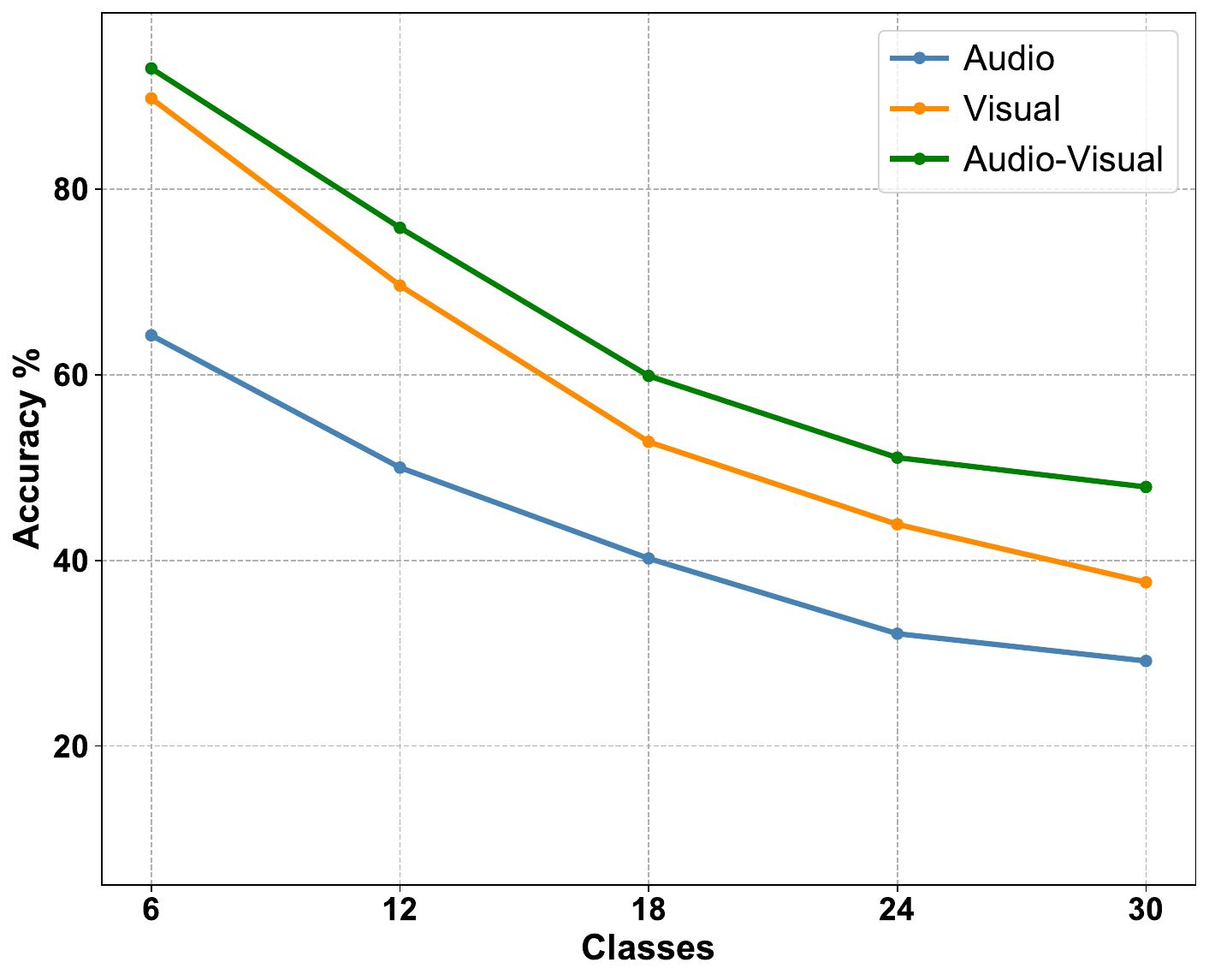}
    \label{fig:acc_KS_iCaRL_FC}
    }
    \\
    \subfloat[]{
    \centering
    \includegraphics[width=0.24\textwidth]{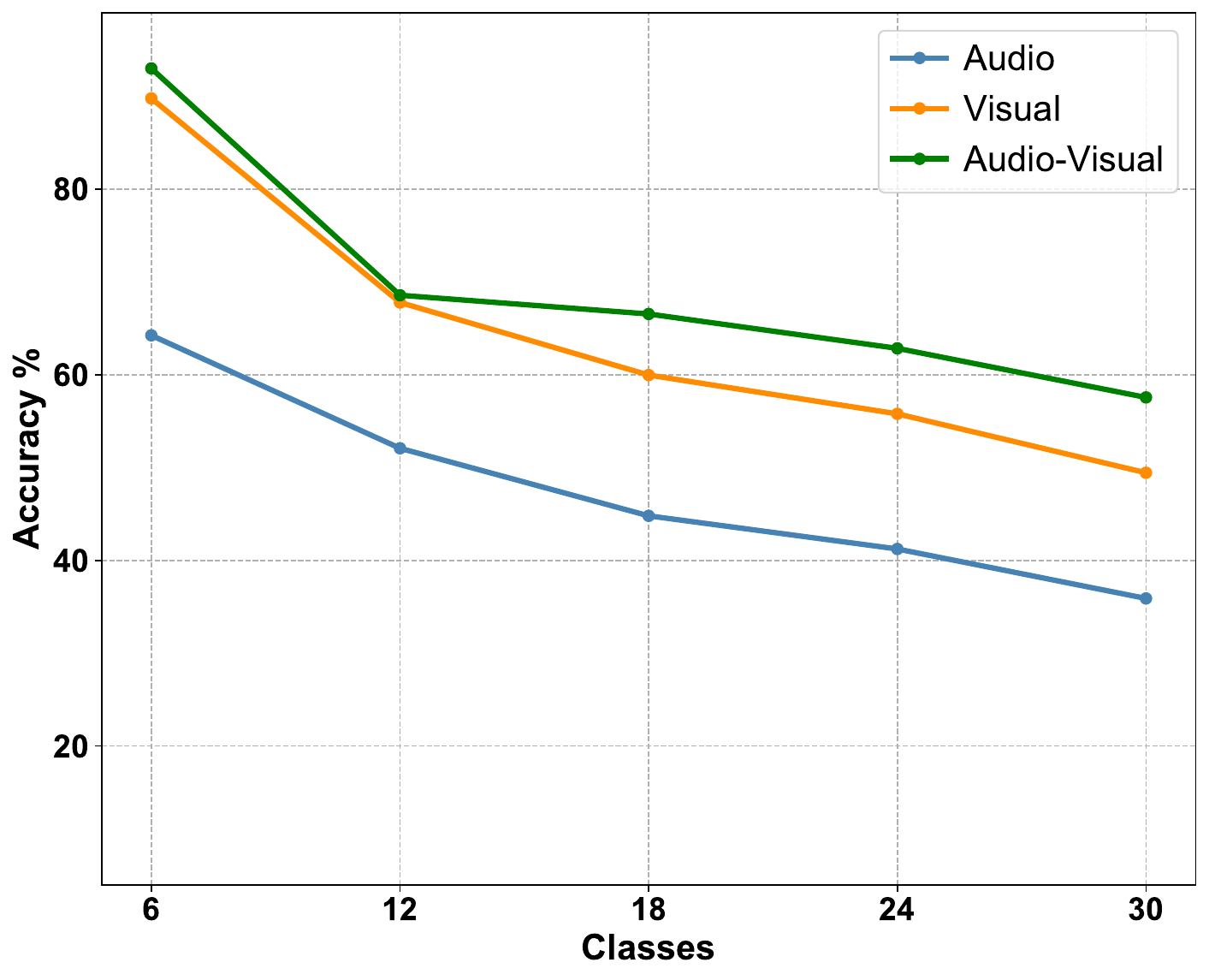}
    \label{fig:acc_KS_SSIL}
    }
    \hspace{-6pt}
    \subfloat[]{
    \centering
    \includegraphics[width=0.24\textwidth]{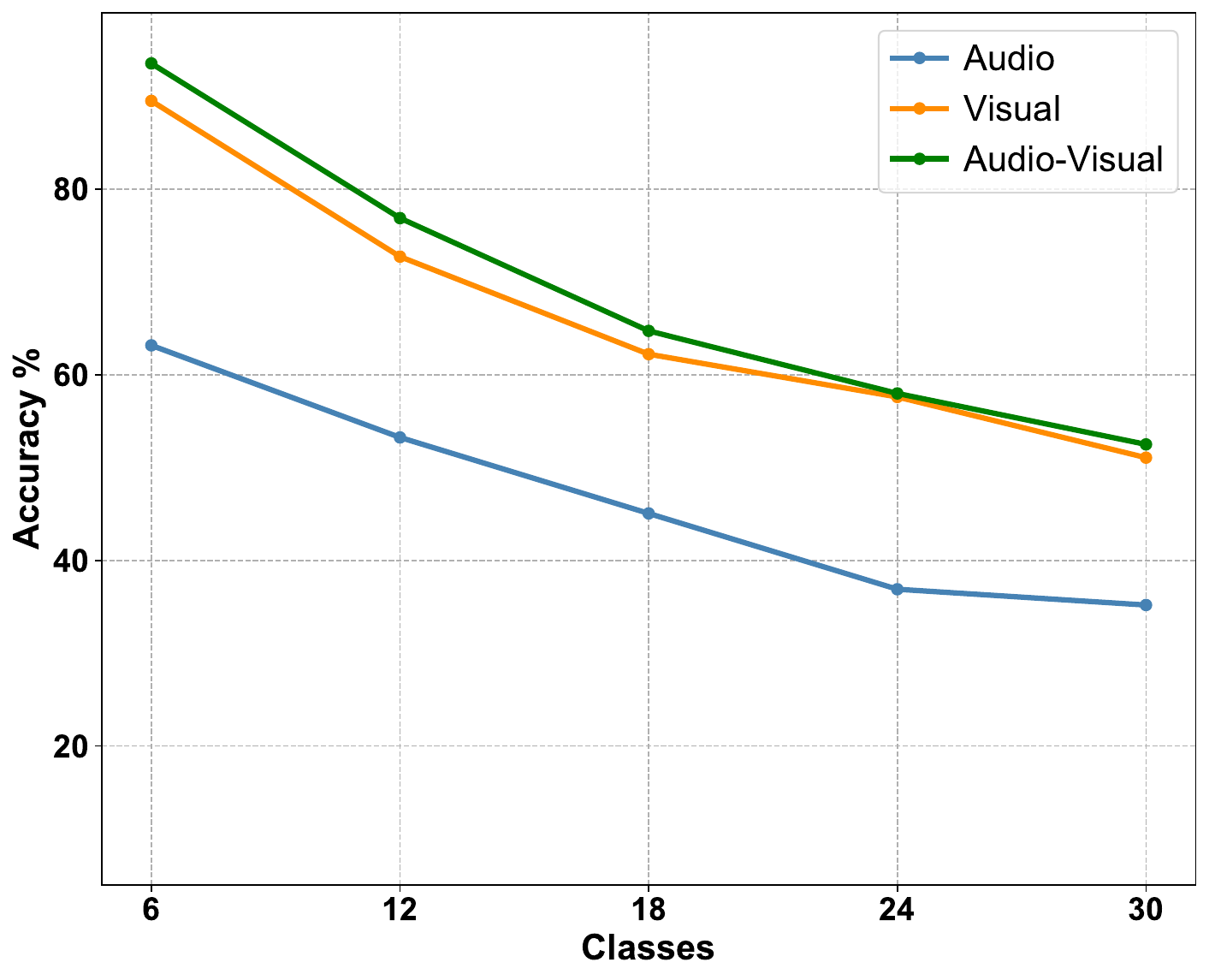}
    \label{fig:acc_KS_AFC_NME}
    }
    \hspace{-6pt}
    \subfloat[]{
    \centering
    \includegraphics[width=0.24\textwidth]{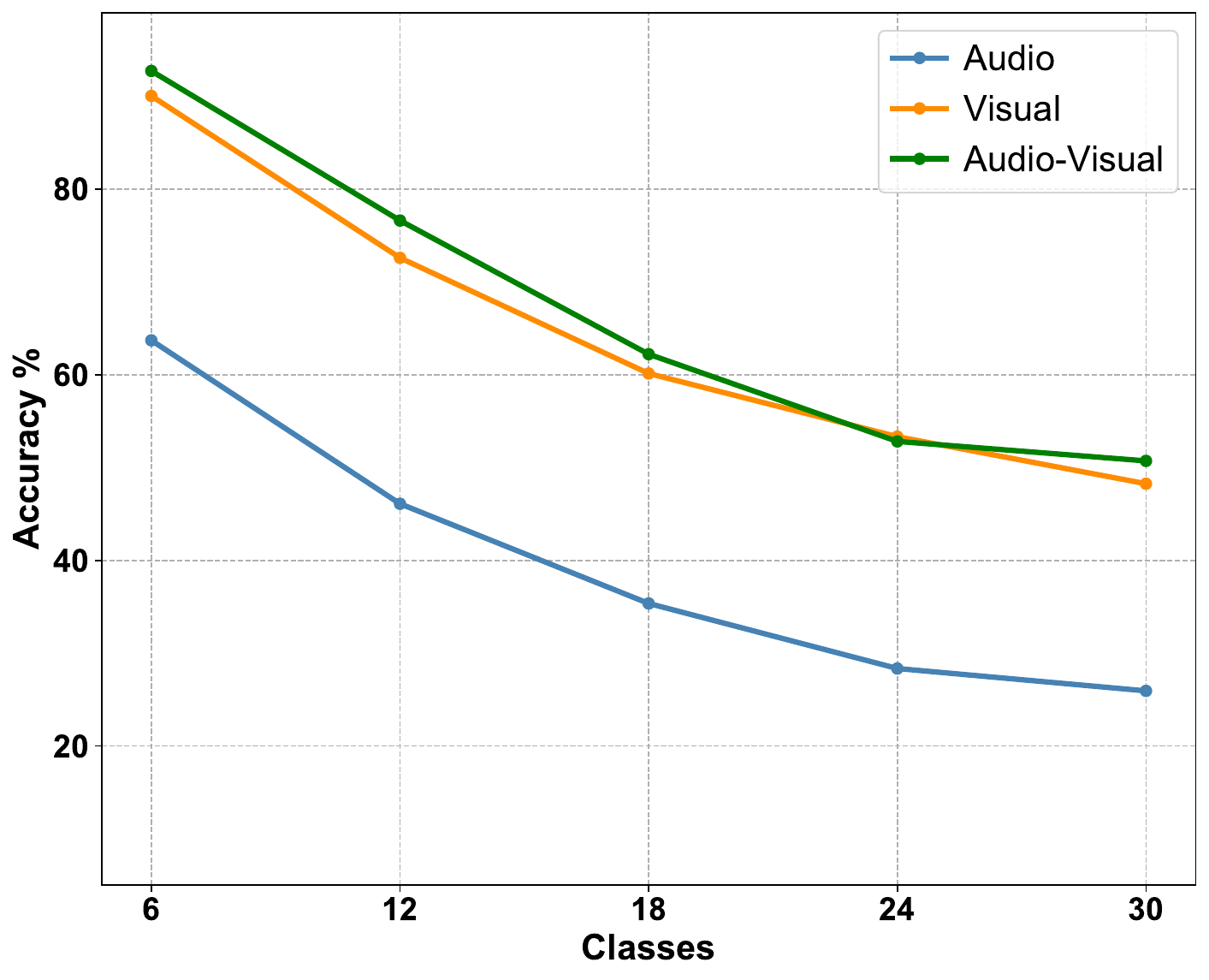}
    \label{fig:acc_KS_AFC_LSC}
    }
    \hspace{-6pt}
    \subfloat[]{
    \centering
    \includegraphics[width=0.24\textwidth]{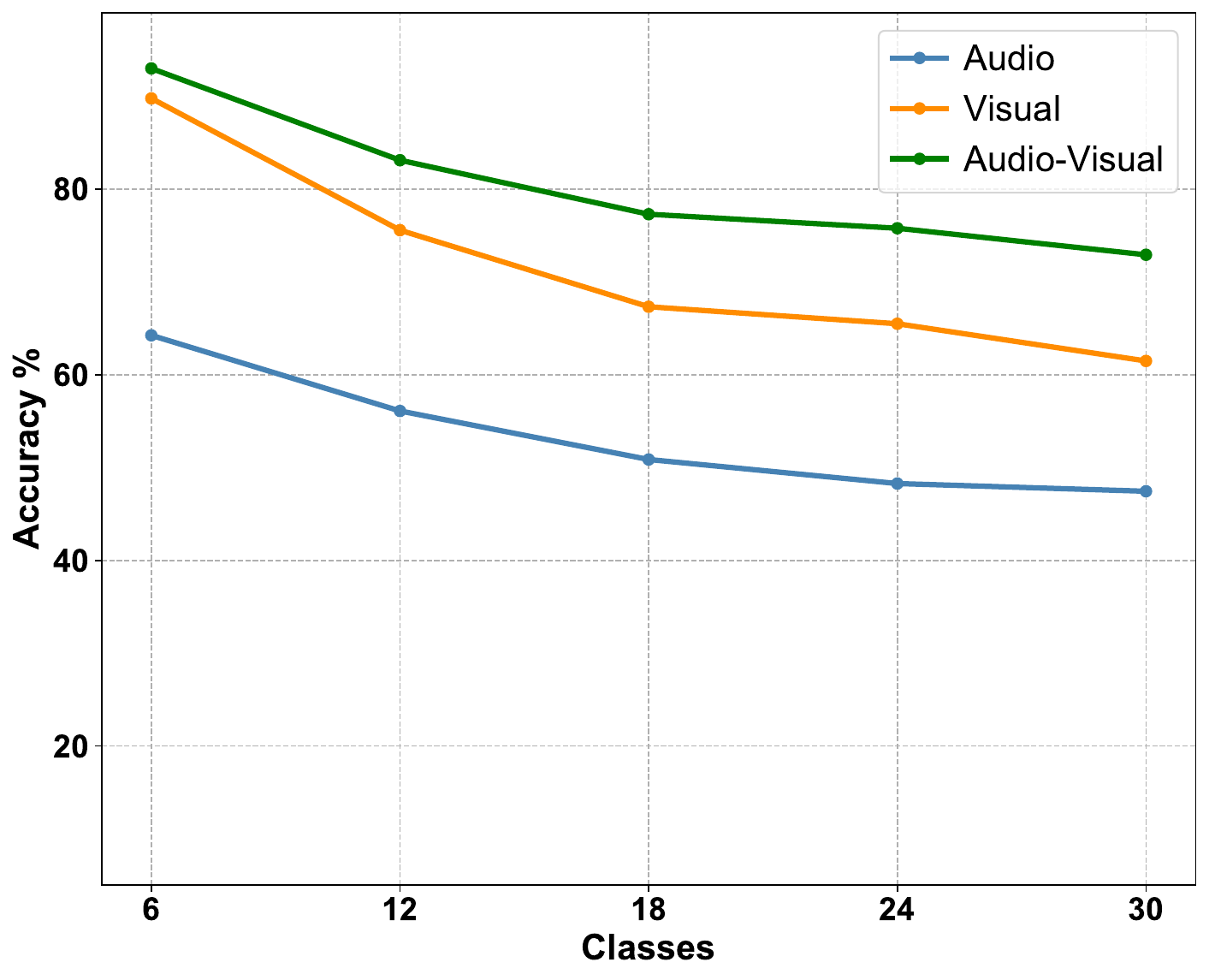}
    \label{fig:acc_KS_upper}
    }
    \caption{Testing accuracy of training with audio, visual and audio-visual modalities of (a) Fine-tuning, (b) LwF, (c) iCaRL-NME, (d) iCaRL-FC, (e) SS-IL, (f) AFC-NME, (g) AFC-LSC, and (h) upper bound on K-S-CI dataset. 
    }
    \label{fig:acc_each_step_baselines_KS}
\end{figure*}

\begin{figure*}[t]
    \centering
    \subfloat[]{
    \centering
    \includegraphics[width=0.24\textwidth]{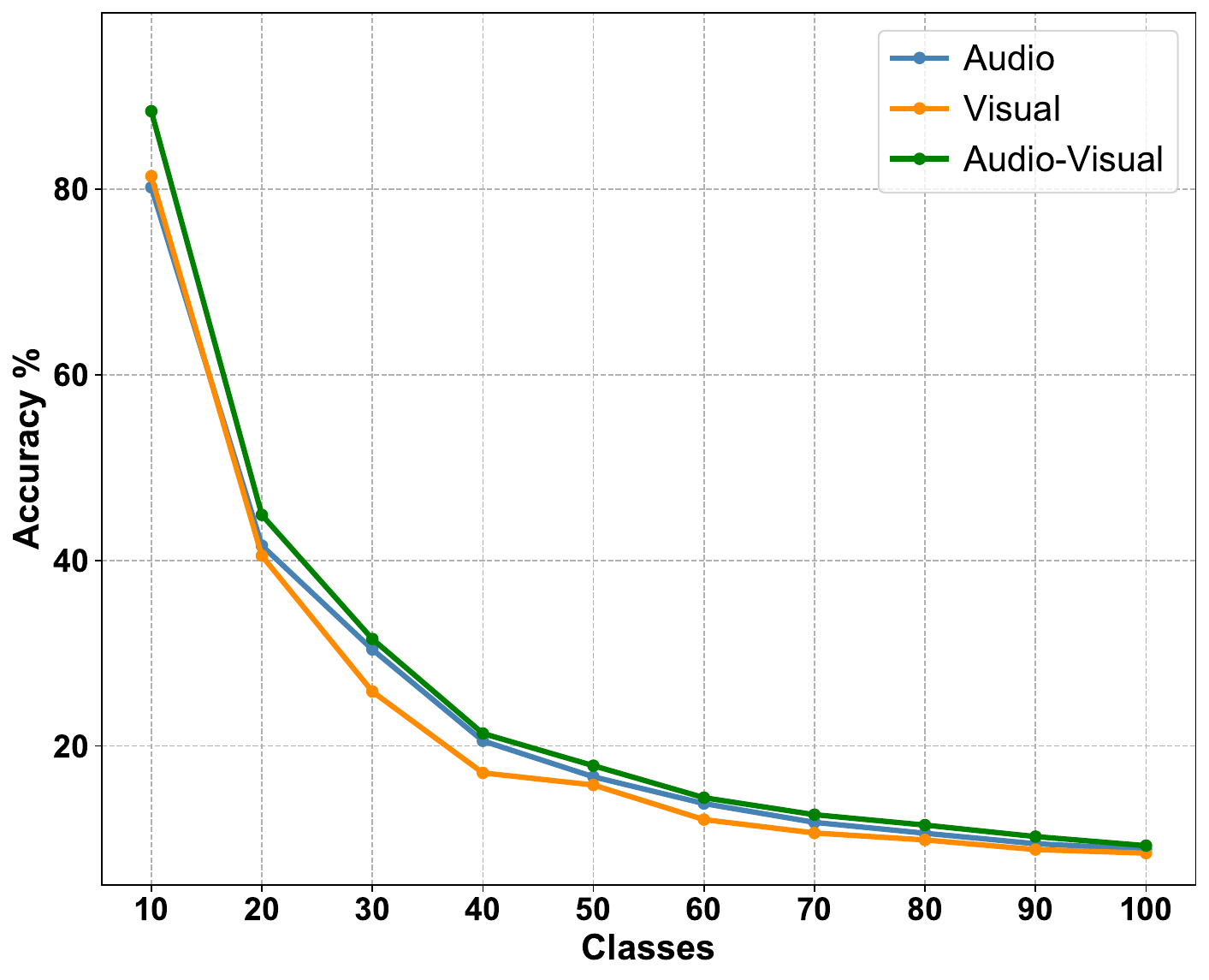}
    \label{fig:acc_VS_fine_tuning}
    }
    \hspace{-6pt}
    \subfloat[]{
    \centering
    \includegraphics[width=0.24\textwidth]{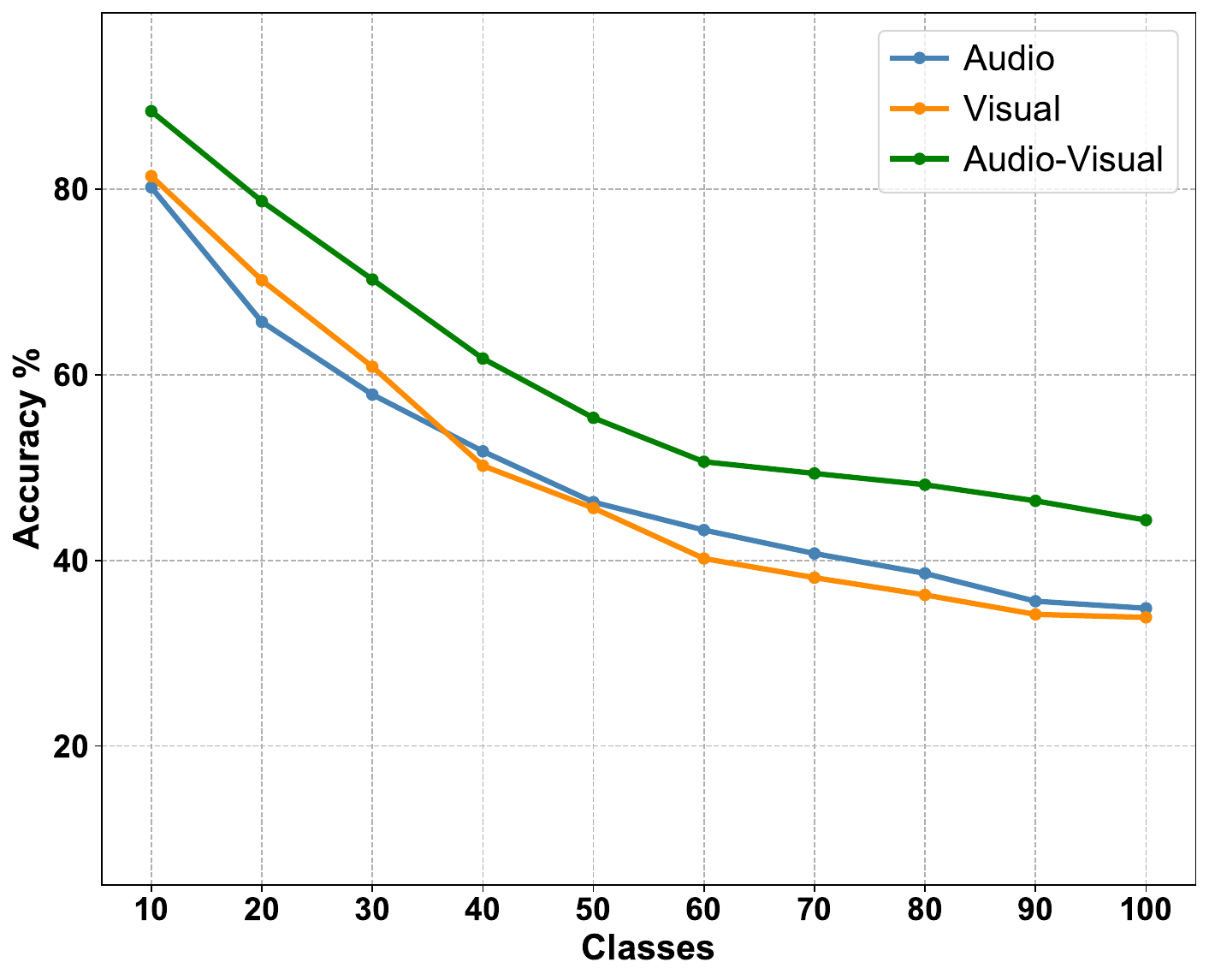}
    \label{fig:acc_VS_lwf}
    }
    \hspace{-6pt}
    \subfloat[]{
    \centering
    \includegraphics[width=0.24\textwidth]{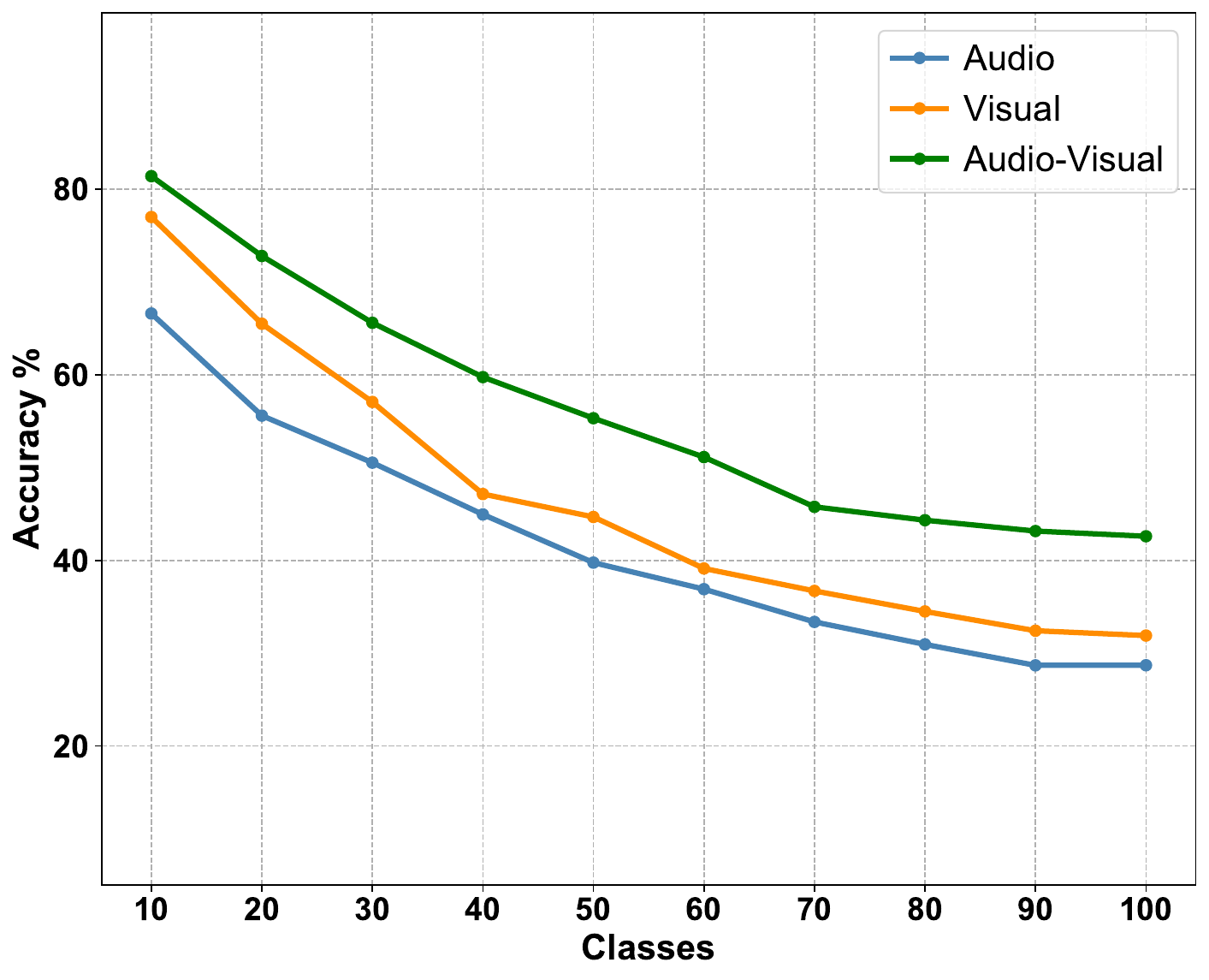}
    \label{fig:acc_VS_iCaRL_NME}
    }
    \hspace{-6pt}
    \subfloat[]{
    \centering
    \includegraphics[width=0.24\textwidth]{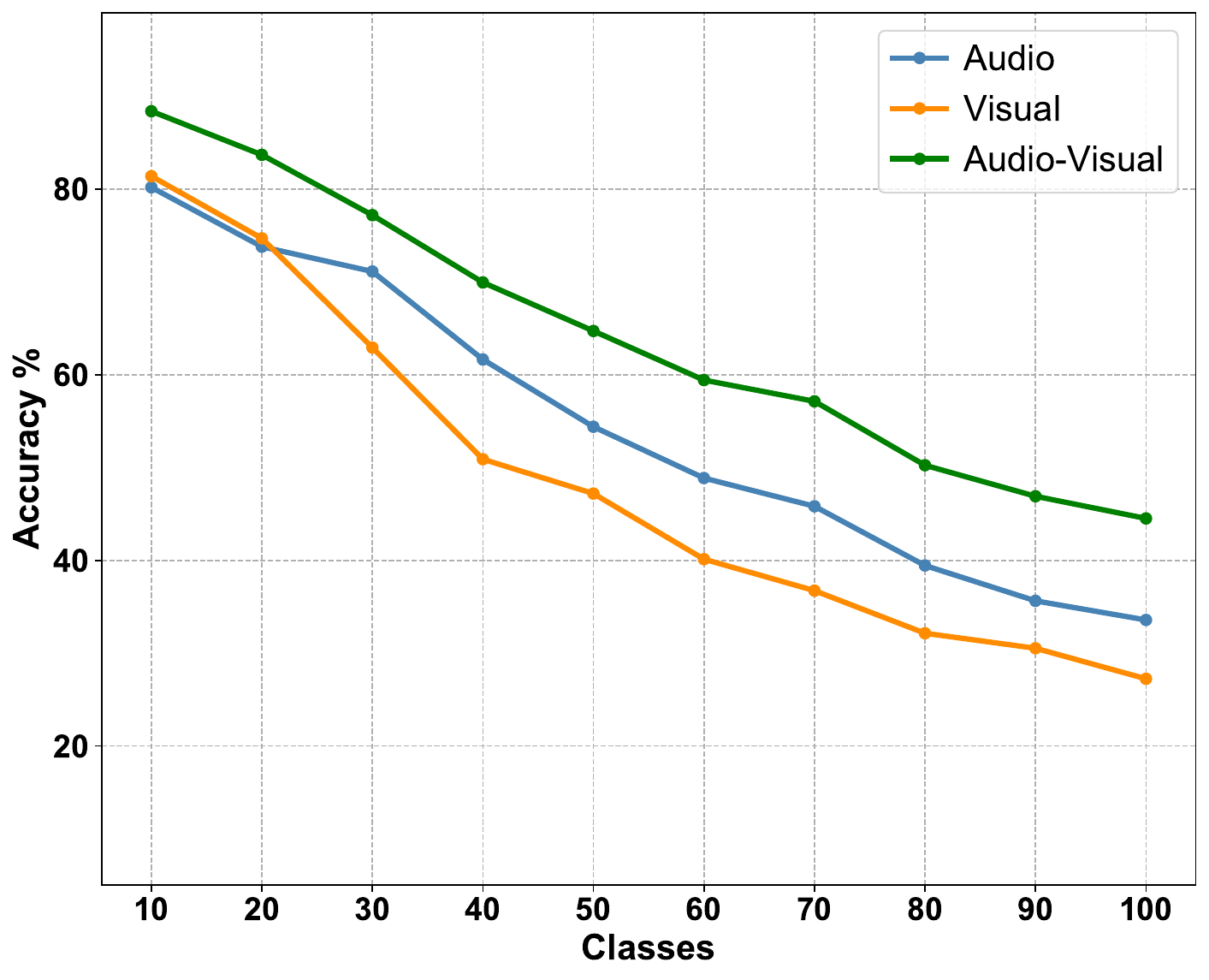}
    \label{fig:acc_VS_iCaRL_FC}
    }
    \\
    \subfloat[]{
    \centering
    \includegraphics[width=0.24\textwidth]{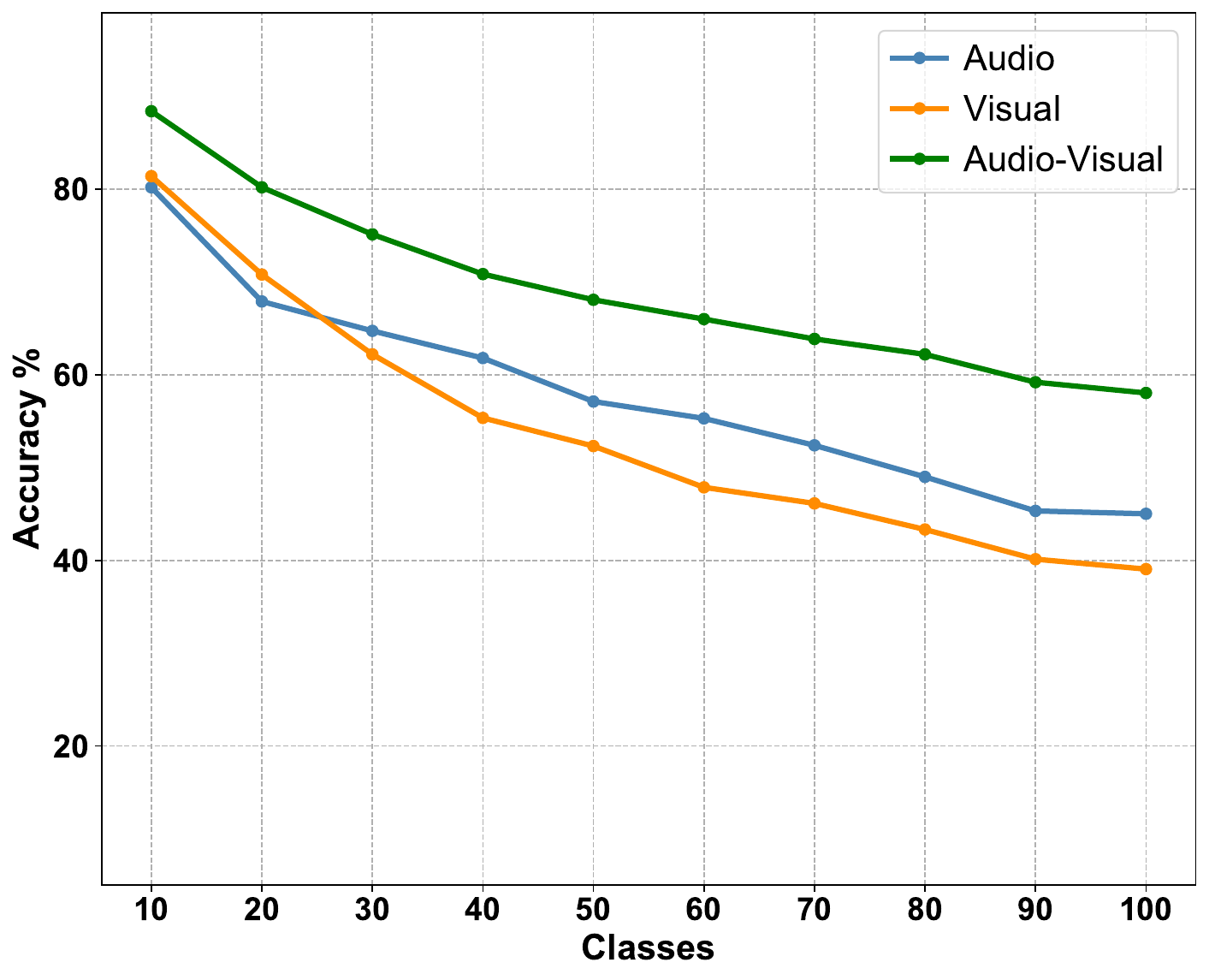}
    \label{fig:acc_VS_SSIL}
    }
    \hspace{-6pt}
    \subfloat[]{
    \centering
    \includegraphics[width=0.24\textwidth]{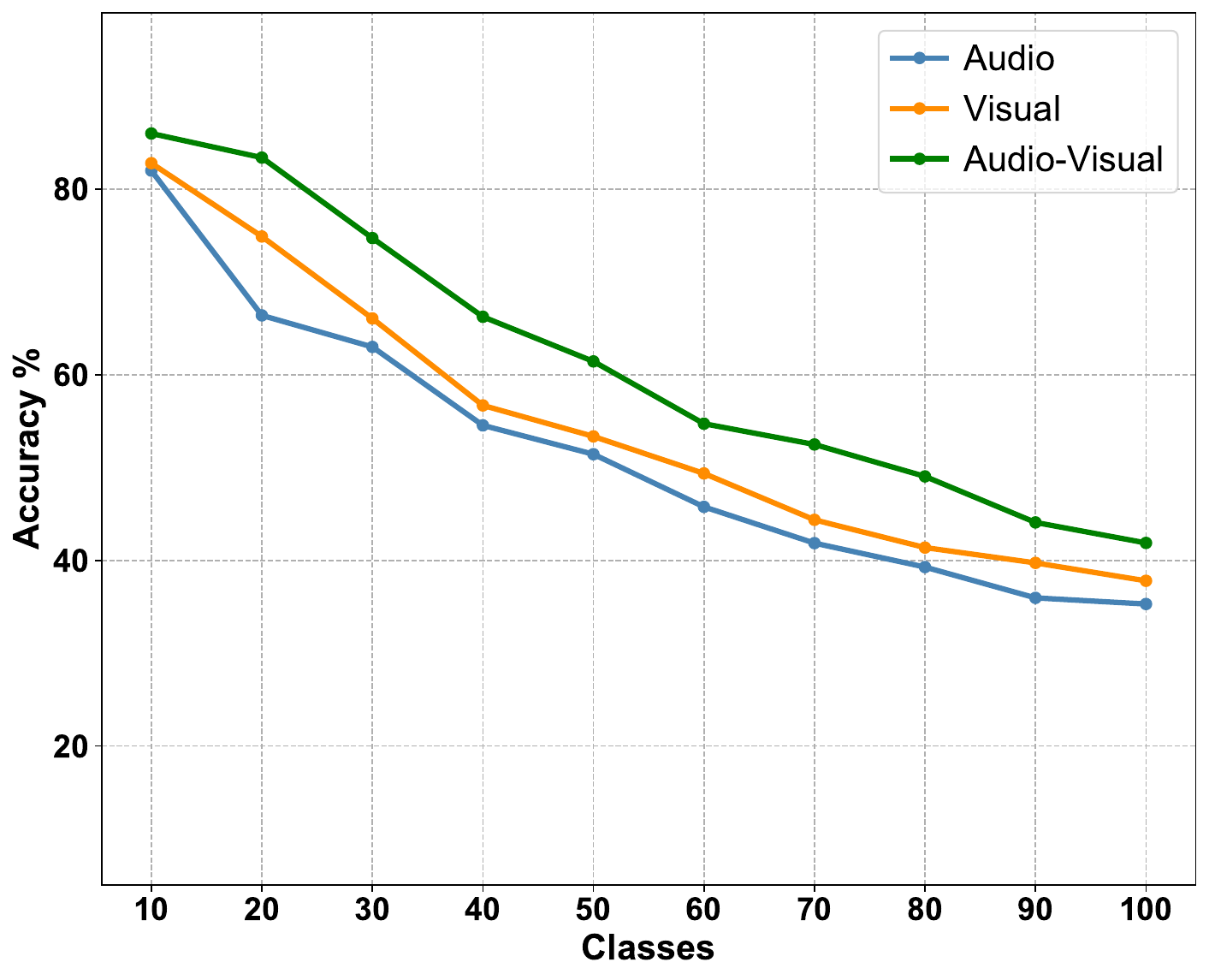}
    \label{fig:acc_VS_AFC_NME}
    }
    \hspace{-6pt}
    \subfloat[]{
    \centering
    \includegraphics[width=0.24\textwidth]{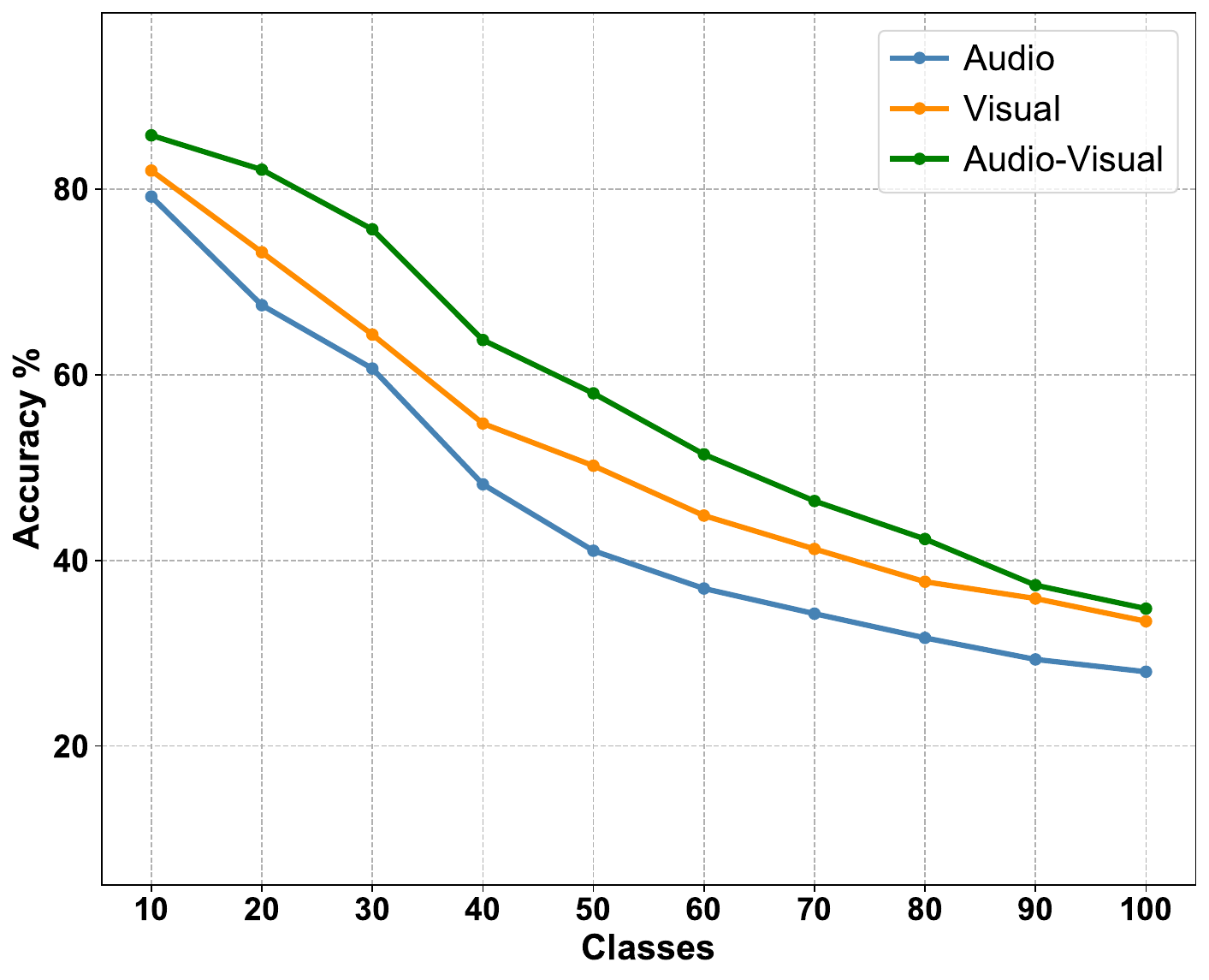}
    \label{fig:acc_VS_AFC_LSC}
    }
    \hspace{-6pt}
    \subfloat[]{
    \centering
    \includegraphics[width=0.24\textwidth]{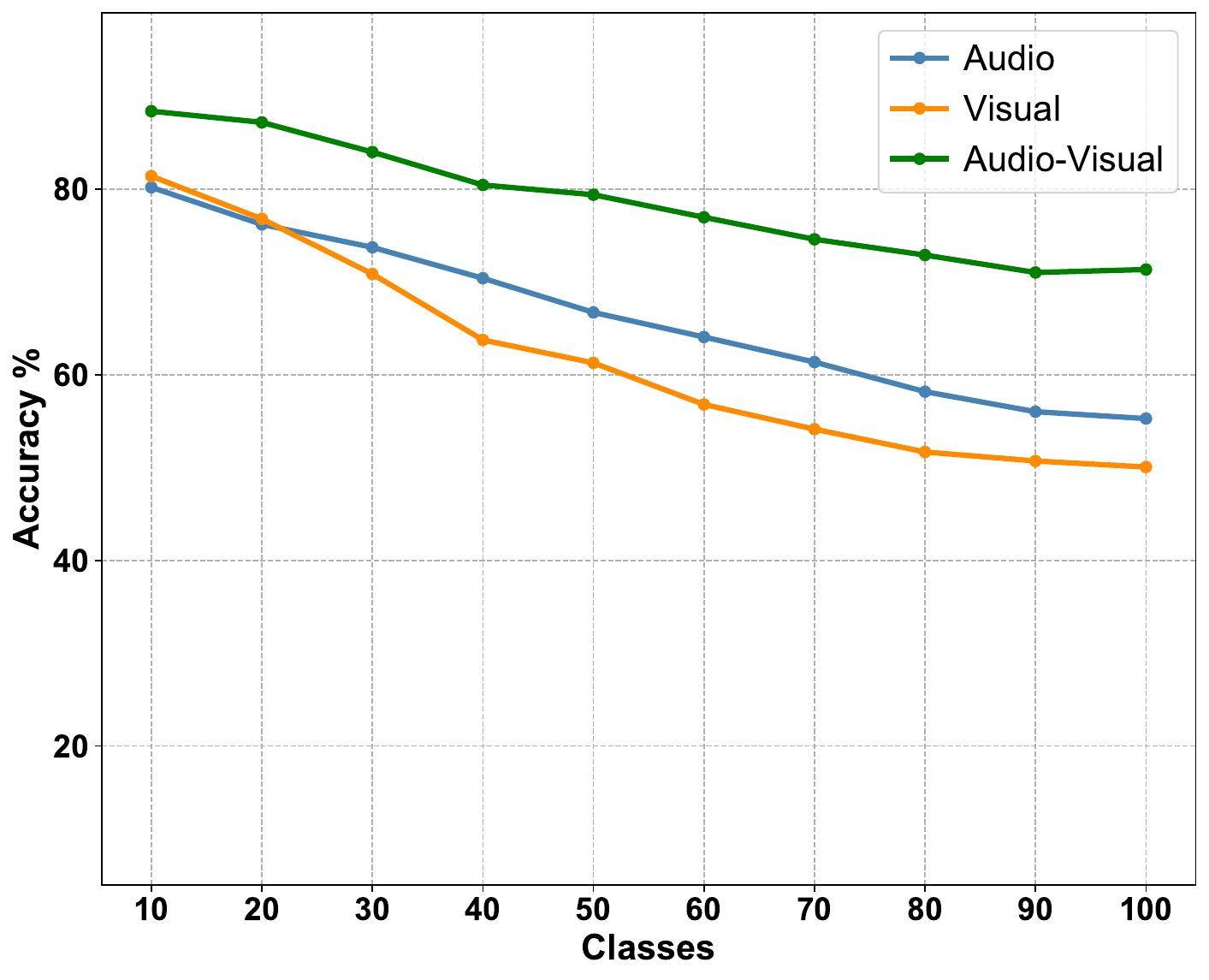}
    \label{fig:acc_VS_upper}
    }
    \caption{Testing accuracy of training with audio, visual and audio-visual modalities of (a) Fine-tuning, (b) LwF, (c) iCaRL-NME, (d) iCaRL-FC, (e) SS-IL, (f) AFC-NME, (g) AFC-LSC, and (h) upper bound on VS100-CI dataset. 
    }
    \label{fig:acc_each_step_baselines_VS}
\end{figure*}

In this section, we present the experimental comparison of several class-incremental learning approaches, including Fine-tuning, LwF~\cite{li2017learning}, iCaRL-NME~\cite{rebuffi2017icarl}, iCaRL-FC~\cite{rebuffi2017icarl}, SSIL~\cite{ahn2021ssil}, AFC-NME~\cite{kang2022class}, and AFC-LSC~\cite{kang2022class}. We compare the performance of these approaches at each step in the context of audio-visual class-incremental learning and unimodal audio or visual modality class-incremental learning.
The experimental results on the AVE-CI, K-S-CI, and VS100-CI datasets are illustrated in Figure~\ref{fig:acc_each_step_baselines_AVE}, Figure~\ref{fig:acc_each_step_baselines_KS}, and Figure~\ref{fig:acc_each_step_baselines_VS}, respectively. These results indicate that training with joint audio-visual modalities consistently outperforms training using only a single audio or visual modality at each incremental step. This highlights the effectiveness and superiority of cross-modal audio-visual learning in class-incremental scenarios compared to single audio or visual modality learning. Overall, our findings suggest that integrating audio and visual modalities in class-incremental learning can enhance performance, promote efficient knowledge transfer between tasks, and prevent the model from forgetting previously acquired knowledge.

\subsection{Full Visualization of Visual Attention}
\label{sec:att}

\begin{figure*}[t]
    \centering
    \includegraphics[width=0.98\textwidth]{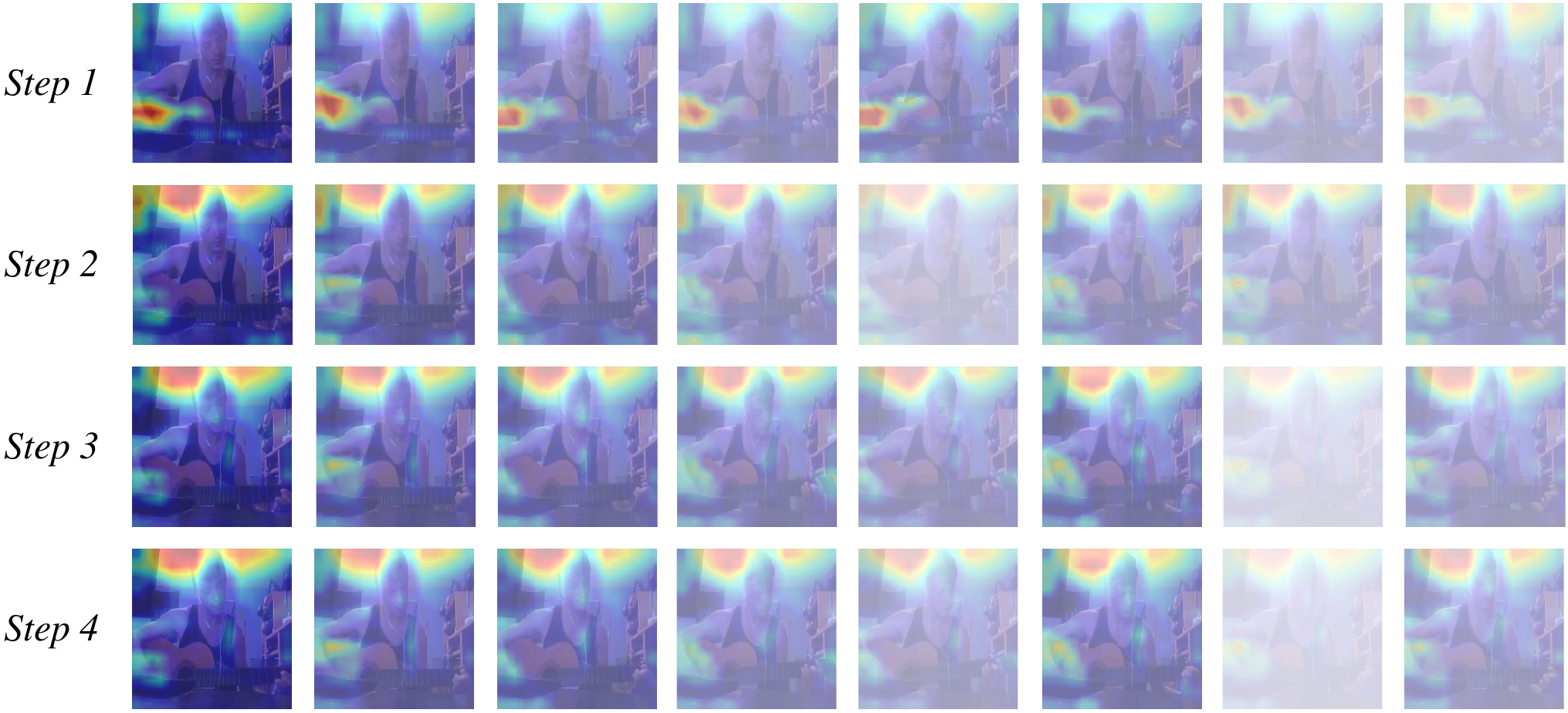}
    \caption{Full version of the visualization of the vanishing of audio-guided visual attention as the incremental step grows.}
    \label{fig:attn_distill_motivation_full}
\end{figure*}

\begin{figure*}[t]
    \centering
    \includegraphics[width=0.98\textwidth]{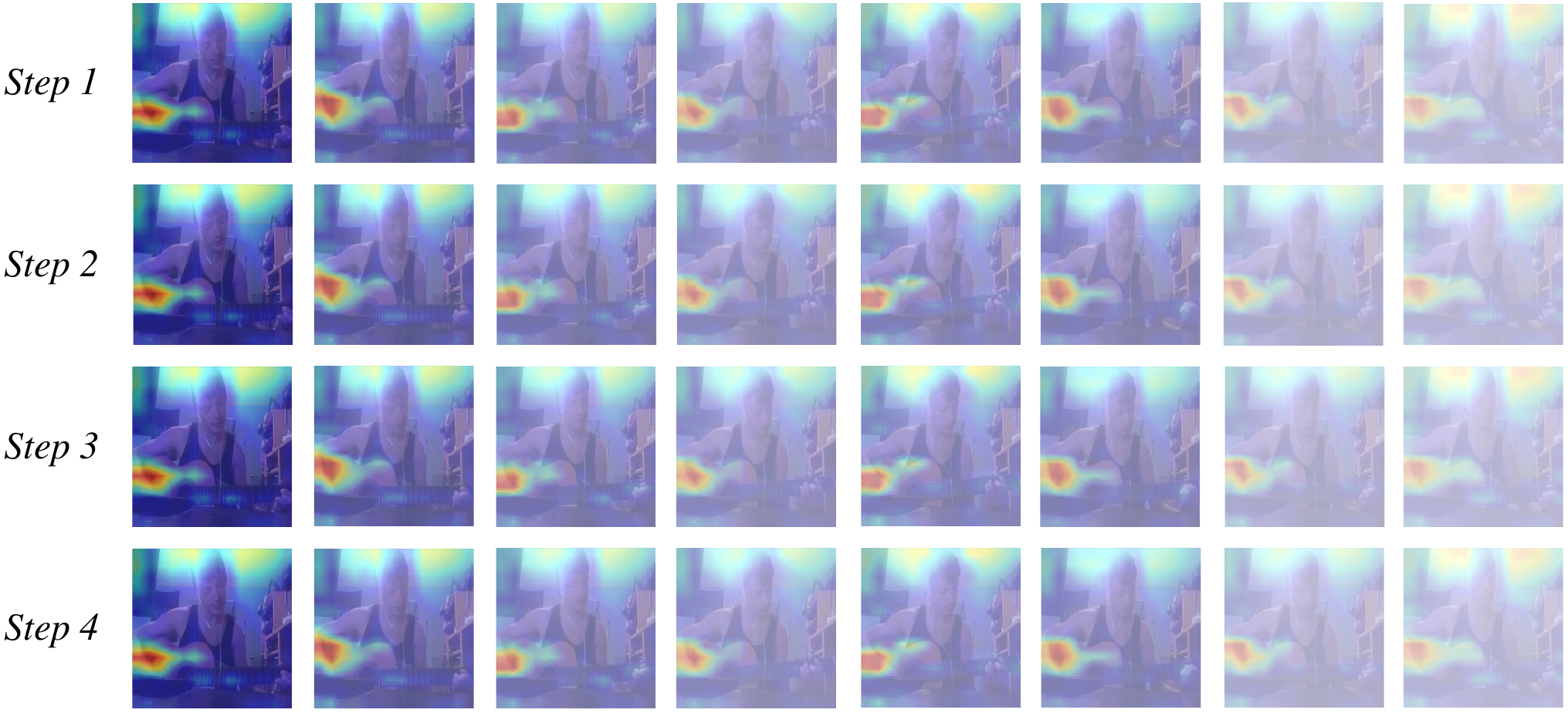}
    \caption{Full version of the visualization of visual attention maps as the incremental step grows with our proposed VAD.}
    \label{fig:attn_distill_after_full}
\end{figure*}

In this section, we present the complete visualization of audio-guided visual attention maps with and without our proposed Visual Attention Distillation (VAD) to demonstrate their vanishing and preservation at each incremental step.
Figure~\ref{fig:attn_distill_motivation_full} provides the full version of the figure initially shown in Section 3.4 of the main paper, illustrating the vanishing of the visual attention map. In Figure~\ref{fig:attn_distill_after_full}, we display the full version of the figure initially shown in Section 4.5, which visualizes the visual attention map after applying our proposed VAD.
These visualizations reveal that implementing our VAD effectively preserves the previously learned audio-guided visual attention capabilities, preventing the model from forgetting the established audio-visual correlations in prior incremental steps. This further substantiates the effectiveness of our proposed approach in averting catastrophic forgetting and maintaining performance throughout multiple incremental learning steps.


\subsection{Parameter Studies}
\label{sec:param}
In this section, we explore the impact of different settings of $\lambda_I$ and $\lambda_C$ on AVE-CI and K-S-CI datasets and present all of our experimental results. We also investigate the impact of $\lambda_{VAD}$ and find that setting it to 0.5 performs well on both datasets, so we use this value in all our experiments. Our experimental results, which can be found in Table~\ref{tab:parameter_study_AVE}, \ref{tab:parameter_study_KS}, and~\ref{tab:parameter_study_VS} for AVE-CI, K-S-CI, and VS100-CI datasets respectively, demonstrate the effects of different settings of $\lambda_I$ and $\lambda_C$ on the performance of the model.

\subsection{Results with Different Incremental Settings}
\label{sec:setting}
In our main experiments, the class-incremental settings of AVE-CI and K-S-CI datasets are set to \textbf{7 classes $\times$ 4 steps} and \textbf{6 classes $\times$ 5 steps} respectively. In this subsection, we conduct experiments on AVE-CI and K-S-CI datasets with the class-incremental settings of \textbf{4 classes $\times$ 7 steps} and \textbf{5 classes $\times$ 6 steps} respectively, to investigate the performance of our proposed method compared to baselines with different class-incremental settings. The results are shown in Table~\ref{tab:diff_step_setting}, from which we can see that, in the incremental setting of \textbf{4 classes $\times$ 7 steps} on AVE-CI dataset, our method outperforms the state-of-the-art method AFC-NME by \textbf{3.15}. For the incremental setting of \textbf{5 classes $\times$ 6 steps} on K-S-CI dataset, our method outperforms state-of-the-art result by \textbf{4.86}. These results demonstrate the versatility of our proposed method, as it can achieve superior performance across different class-incremental settings.

\subsection{Comparison with Existing Visual Attention Distillation Approach}
\label{sec:visual_attn_distil}
\begin{figure}
    \centering
    \includegraphics[width=0.45\textwidth]{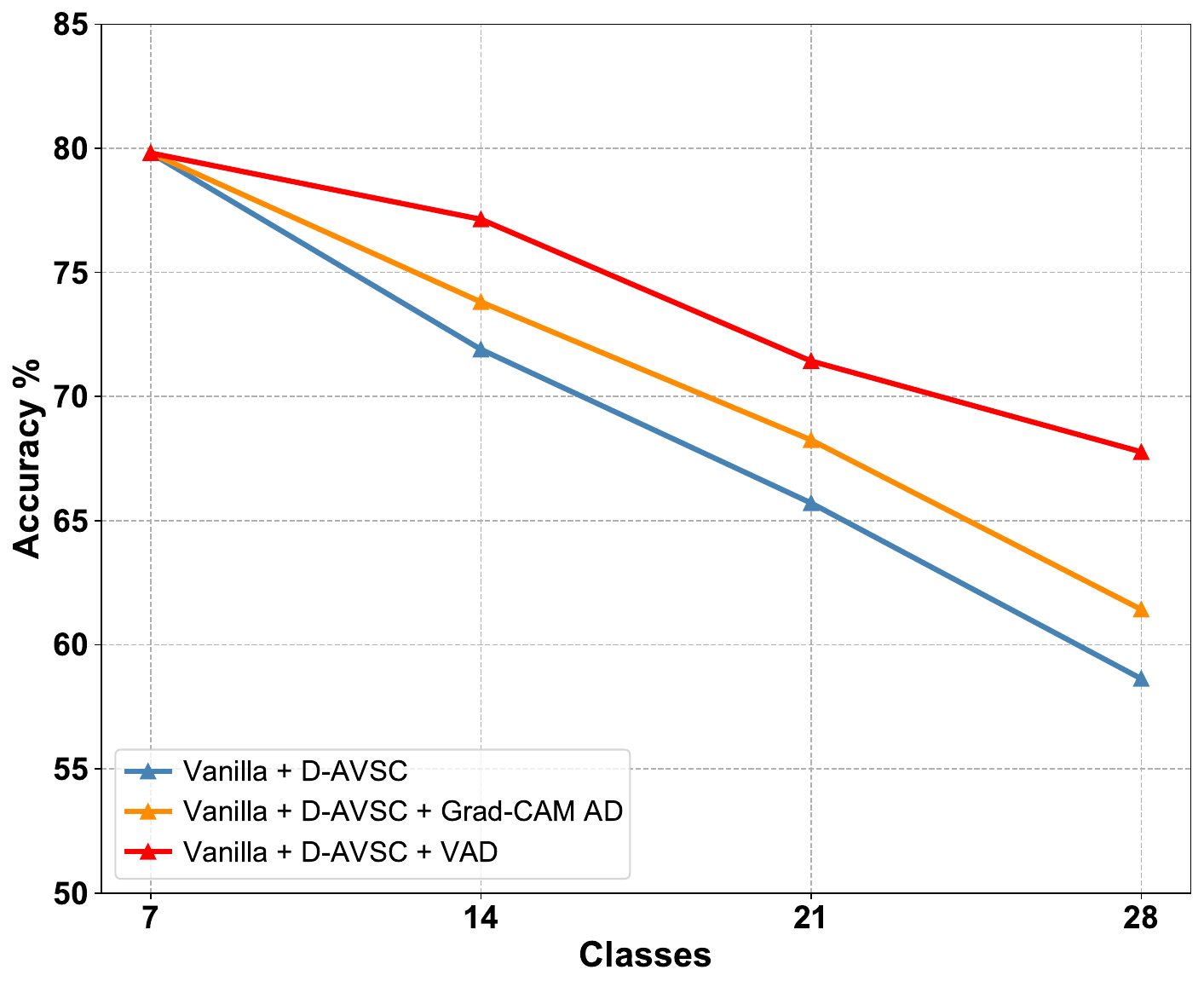}
    \caption{Testing accuracy at each incremental step of (1) Vanilla + D-AVSC, (2) Vanilla + D-AVSC + Grad-CAM AD, and (3) Vanilla + D-AVSC + VAD on AVE-CI dataset.}
    \label{fig:acc_variants_grad_cam}
\end{figure}

In our AV-CIL, we propose the VAD, a novel audio-guided visual attention map distillation method for audio-visual class-incremental learning. Our VAD enablea the model to preserve previously learned attentive ability in future classes/tasks, effectively preventing the model from forgetting previously learned audio-visual semantic correlations. To further evaluate the effectiveness of our proposed VAD compared to existing attention distillation methods, we construct a variant of our AV-CIL, by replacing our VAD with the Grad-CAM-based visual attention distillation (Grad-CAM AD)~\cite{dhar2019learning}. We conduct experiments on the AVE-CI dataset with different variants of our proposed AV-CIL, in which we name Vanilla as the variant without the D-AVSC and the VAD (with only the Task-wise Knowledge Distillation and the Separated Softmax Cross-Entropy). The experimental results are presented in Table~\ref{tab:attn_dist_variants}, where we show the Mean Accuracy of Vanilla + D-AVSC, Vanilla + D-AVSC + Grad-CAM AD, and Vanilla + D-AVSC + VAD on AVE-CI dataset. From the table, we can see that our full AV-CIL outperforms the variant with Grad-CAM AD significantly, which demonstrates the superiority and effectiveness of our proposed VAD over the Grad-CAM AD. We also show the testing accuracy at each incremental step in Figure~\ref{fig:acc_variants_grad_cam}, where we can see that our full AV-CIL has better performance at each incremental step compared to the variants with Grad-CAM AD, further demonstrating the effectiveness of our proposed VAD.

\begin{table}[htbp]
  \centering
  \caption{Experimental results of different variants on AVE-CI dataset. Our AV-CIL performs better than the variant with Grad-CAM AD, demonstrating the superiority of our proposed VAD over the Grad-CAM AD.}
    \begin{tabular}{lc}
    \toprule
    Variants & Mean Acc. \\
    \midrule
    Vanilla + D-AVSC & 69.01 \\
    Vanilla + D-AVSC + Grad-CAM AD & 70.82 \\
    Vanilla + D-AVSC + VAD (AV-CIL) & \textbf{74.04} \\
    \bottomrule
    \end{tabular}%
  \label{tab:attn_dist_variants}%
\end{table}%

\end{document}